\documentclass[preprint,12pt]{elsarticle}

\usepackage{amssymb}
\usepackage{algorithm}
\usepackage{algpseudocode}
\usepackage{booktabs}
\usepackage{siunitx} 
\usepackage{natbib}
\usepackage{subcaption} 
\usepackage{fontspec}
\usepackage[colorlinks=true, linkcolor=green, urlcolor=green, citecolor=green]{hyperref}
\usepackage[table]{xcolor} 

\usepackage{multirow}
\usepackage{bbding}
\usepackage{amsfonts}
\usepackage{makecell}

\usepackage[symbol]{footmisc} 
\usepackage{amsmath}

\newcommand{\totalframework}{MADCrowner}
\newcommand{\marginseg}{CrownSegger}
\newcommand{\crowngen}{CrownDeformR}

\journal{Medical Image Analysis}

\begin{document}

\begin{frontmatter}


\fntext[pjl]{Linda Wei and Chang Liu are co-first authors}
\cortext[cor1]{Corresponding authors}

\title{MADCrowner: Margin Aware Dental Crown Design with Template Deformation and Refinement}

\author[label1]{Linda Wei \fnref{pjl}}
\author[label2,label3]{Chang Liu \fnref{pjl}}
\author[label4]{Wenran Zhang}
\author[label1]{Yuxuan Hu}
\author[label5]{Ruiyang Li}
\author[label8]{Feng Qi}
\author[label1]{Changyao Tian}
\author[label1]{Ke Wang}
\author[label2]{Yuanyuan Wang}

\author[label3]{Shaoting Zhang \corref{cor1}}
\author[label6]{Dimitris Metaxas}
\author[label1,label7]{Hongsheng Li \corref{cor1}}

\affiliation[label1]{organization={Multimedia Laboratory, The Chinese University of Hong Kong},
            postcode={999077},
            state={Hong Kong SAR},
            country={China}}

\affiliation[label2]{organization={College of Biomedical Engineering, Fudan University},
            postcode={200433},
            state={Shanghai},
            country={China}}

\affiliation[label3]{organization={Sensetime Research},
            postcode={200233},
            state={Shanghai},
            country={China}}
            
\affiliation[label4]{organization={Department of Second Dental Center, Shanghai Ninth People's Hospital, Shanghai Jiao Tong University School of Medicine},
            postcode={200011},
            state={Shanghai},
            country={China}}

\affiliation[label5]{organization={Department of Computer Science and Engineering, The Chinese University of Hong Kong},
            postcode={999077},
            state={Hong Kong SAR},
            country={China}}

\affiliation[label6]{organization={Department of Computer Science, Rutgers University},
            postcode={08854},
            state={New Jersey},
            country={USA}}

\affiliation[label7]{organization={Centre for Perceptual and Interactive Intelligence (CPII) under InnoHK},
            postcode={999077},
            state={Hong Kong SAR},
            country={China}}

\affiliation[label8]{organization={Shanghai Stomatological Hospital, Fudan University},
            postcode={200032},
            state={Shanghai},
            country={China}}


\begin{abstract}
Dental crown restoration is one of the most common treatment modalities for tooth defect, where personalized dental crown design is critical. While computer-aided design (CAD) systems have notably enhanced the efficiency of dental crown design, extensive manual adjustments are still required in the clinic workflow. Recent studies have explored the application of learning-based methods for the automated generation of restorative dental crowns. Nevertheless, these approaches were challenged by inadequate spatial resolution, noisy outputs, and overextension of surface reconstruction. To address these limitations, we propose \totalframework, a margin-aware mesh generation framework comprising \crowngen~and \marginseg. Inspired by the clinic manual workflow of dental crown design, we designed \crowngen~to deform an initial template to the target crown based on anatomical context, which is extracted by a multi-scale intraoral scan encoder. Additionally, we introduced \marginseg, a novel margin segmentation network, to extract the cervical margin of the target tooth. The performance of \crowngen~improved with the cervical margin as an extra constraint. And it was also utilized as the boundary condition for the tailored postprocessing method, which removed the overextended area of the reconstructed surface. We constructed a large-scale intraoral scan dataset and performed extensive experiments. The proposed method significantly outperformed existing approaches in both geometric accuracy and clinical feasibility. The code repository was released in \url{https://github.com/lullcant/MADCrowner}.

\end{abstract}

\begin{graphicalabstract}
\centering
\includegraphics[width=\linewidth]{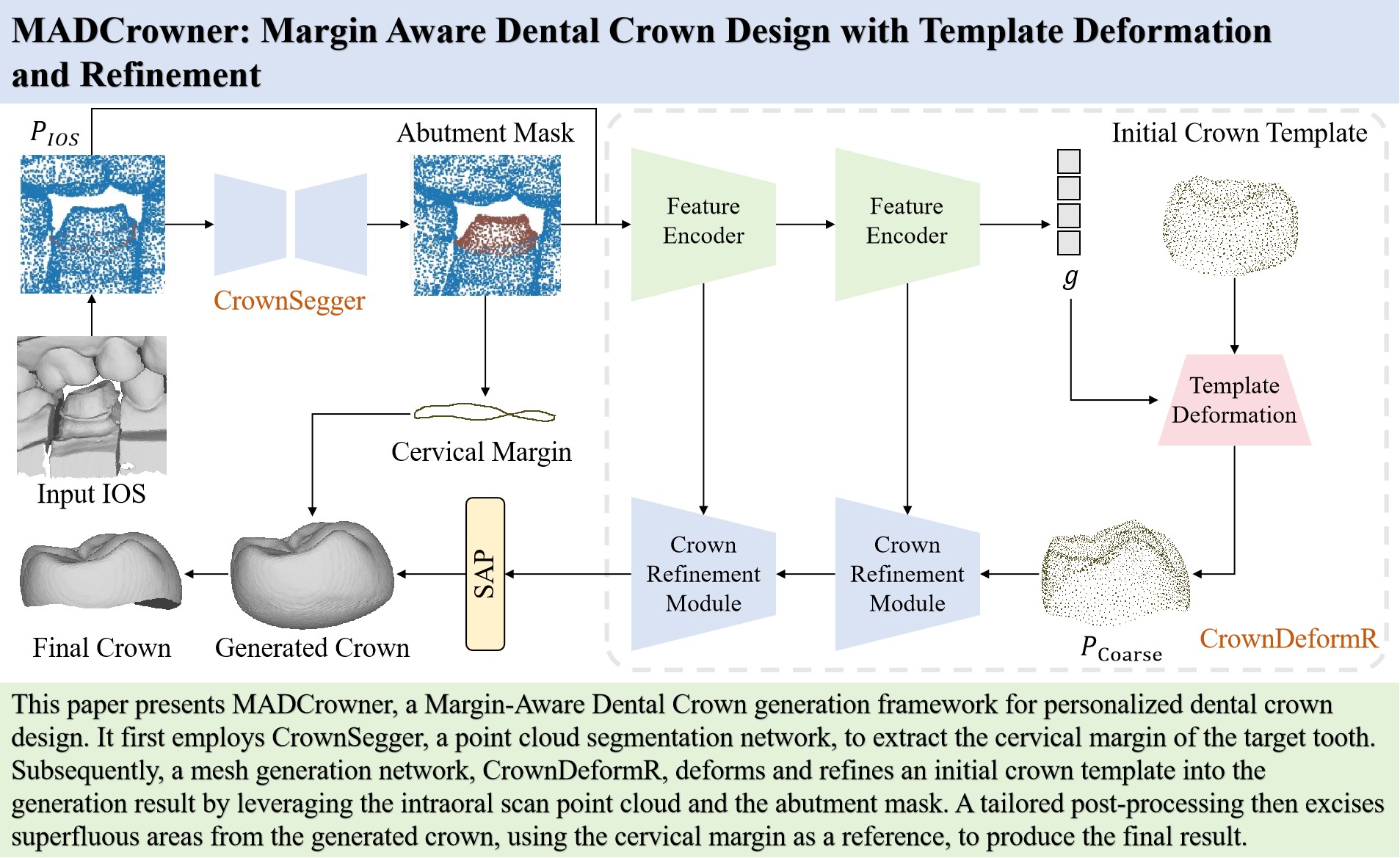}
\end{graphicalabstract}

\begin{highlights}
\item Designed a cervical margin-aware AI framework, \totalframework, for automated dental crown design.

\item Introduced a compact model, \marginseg, to extract the cervical margin.

\item Proposed a novel network, \crowngen, to generate dental crowns from coarse to fine, leveraging anatomical context and cervical margin constraint.

\item Eliminated the superfluous area generated by surface reconstruction by a tailored post-processing method.

\item Experimented on a large-scale clinical intraoral scan dataset and achieved the SOTA performance.
\end{highlights}


\end{frontmatter}




\section{Introduction}
\begin{figure*}
    \centering
    \includegraphics[width=\linewidth]{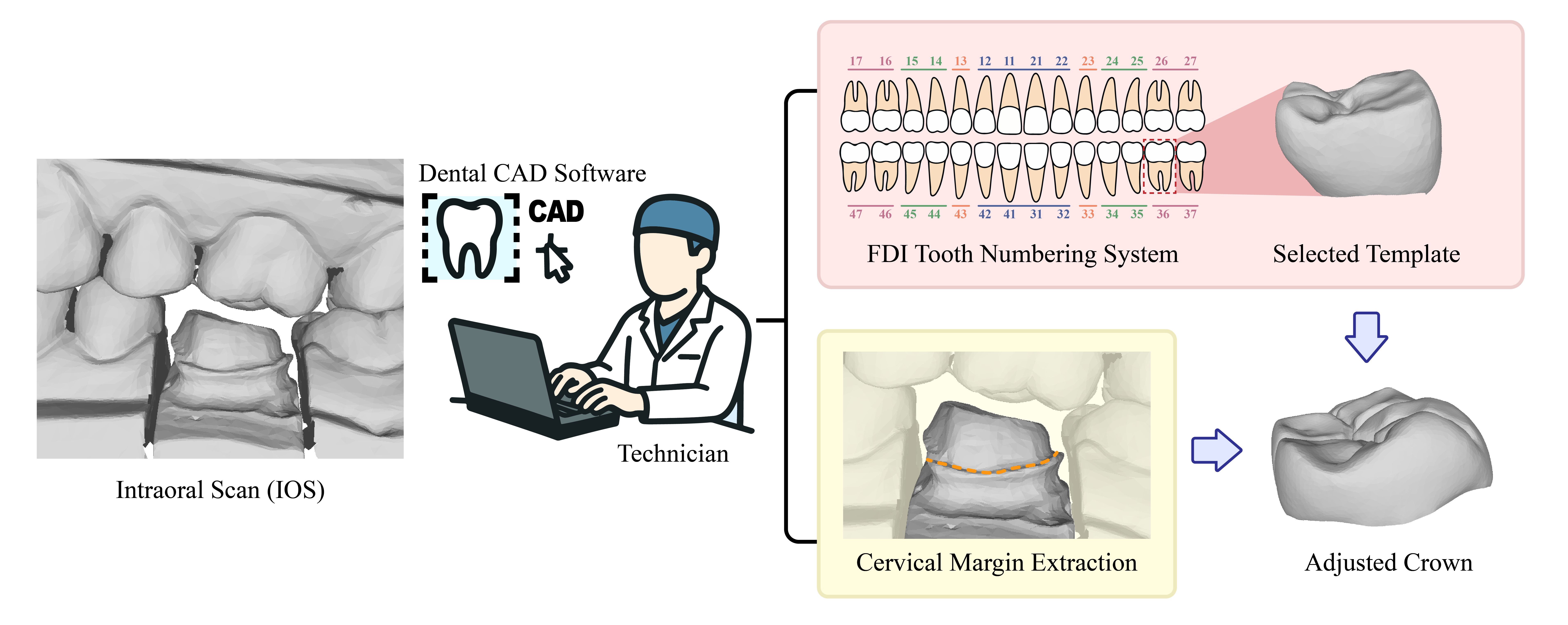}
    \caption{Dental Crown Designing Procedure. With the patient's intraoral scan, the technician utilizes a CAD system to extract the cervical margin, select and modify a digital template of the target tooth, culminating in the design of the final crown.}
    \label{fig:fig1}
\end{figure*}

Oral diseases are among the most prevalent non-communicable diseases globally, impacting approximately 3.5 billion individuals \citep{world2022global,peres2019oral,leo2025impact}. Tooth defect, often originating from caries trauma or gene, is a significant oral health issue~\citep{zaleckiene2014traumatic}, for which crown restoration is extensively utilized and recognized as an effective treatment modality. In clinical dentistry, the procedure for crown restoration typically commences with the preparation of the carious or root canal-treated tooth \citep{hollins2024basic,goodacre2001tooth,podhorsky2015tooth}. The affected tooth was shaped into an abutment to provide adequate retention and resistance for the final prosthesis. Subsequently, a personalized crown is meticulously designed based on the patient's oral environment to achieve optimal restoration for both esthetic and occlusal function. The fabricated crown is then luted onto the prepared abutment, thereby reestablishing the geometry and function of the restored dentition.

The advent of computer-aided design (CAD) systems has significantly simplified crown design \citep{davidowitz2011use,jain2016cad,miyazaki2009review,ardila2023efficacy,rexhepi2023clinical}. As illustrated in \autoref{fig:fig1}, the intraoral scan (IOS) data after abutment preparation are first imported into the CAD system. The technician begins by delineating the cervical margin on the prepared abutment to ensure that the designed crown fits precisely with the intraoral geometry. Subsequently, an appropriate template for the target tooth is selected and adjusted to conform to the morphology and alignment of both the opposing and adjacent dentition, culminating in a customized dental crown. While CAD systems have significantly enhanced the efficiency of dental technicians, a critical challenge remains: the built-in dental crown templates are standardized. This inherent uniformity requires technicians to diligently modify these templates to accommodate the individual intraoral anatomy of each patient. In clinical practice, dental crown design still requires a considerable amount of time, typically ranging from 15 minutes to an hour for a single tooth \citep{bessadet2025time,joda2016time}. Therefore, the development of a framework for the automatic design of personalized crowns has substantial clinical value, as it significantly reduces the workload and time demands for dental technicians.

\begin{figure}
    \centering
    \includegraphics[width=\linewidth]{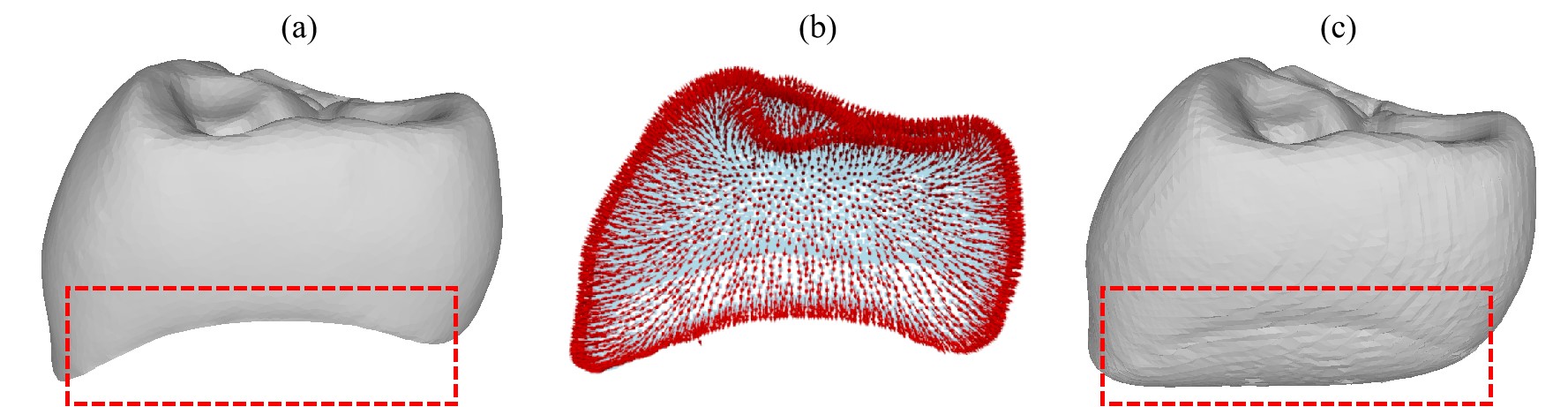}
    \caption{The visualization of surface reconstruction for a certain dental crown. (a) The original crown mesh. (b) point cloud with normal vectors of the crown mesh. (c) Reconstructed surface of (b) by Poisson surface reconstruction. The bottom of the reconstructed surface is compulsorily sealed due to the intrinsic constraints of the reconstruction algorithm.}
    \label{fig:recon}
\end{figure}

With the development of artificial intelligence, deep learning has been increasingly adopted for medical image analysis.~\citep{shen2017deep,wang2024sensecare}. Several studies have leveraged deep learning algorithms to assist in dental crown design. Some of these approaches project IOS data from the occlusal perspective onto 2D depth maps, which are then used with image generation algorithms to create the intricate occlusal surface details of the crown \citep{tian2021dcpr,yang2025mvdc,isola2017image}. However, these methods suffer from an intrinsic limitation: information loss introduced by the projection. Some crucial anatomical regions, such as the cervical margin, are obscured in 2D projections. The 2D depth projection is inherently incomplete, as it lacks data for regions obscured from the occlusal view. Post-processing is required to complete these areas. To address these issues, some recent studies have attempted to utilize point cloud completion networks combined with surface reconstruction algorithms for crown generation \citep{hosseinimanesh2025personalized,hosseinimanesh2023mesh,yang2024dcrownformer}. Nonetheless, such approaches increase the learning complexity, and there is an inherent trade-off between minimizing output noise and enhancing the details of the generation results. Moreover, the surface reconstruction algorithm in previous studies inherently produces watertight meshes \citep{lorensen1998marching,peng2021shape}. This process is fundamentally incompatible with a dental crown, which is an open genus-zero mesh. As illustrated in \autoref{fig:recon}, there is a noticeable overextension at the bottom of the reconstructed crown surface. These limitations constrain the clinical applicability of crown generation algorithms.

In this work, we proposed a \textbf{M}argin \textbf{A}ware \textbf{D}ental \textbf{Crown} generation framework (\textbf{\totalframework}) for personalized dental crown design, primarily composed of two modules: \marginseg~and \crowngen. The \marginseg~combines both point cloud and volumetric information to segment the prepared abutment region from IOS data, thus accurately determining the cervical margin of the target abutment. In \crowngen, the output of \marginseg~and the IOS data are integrated as anatomical context features via a multiscale point cloud encoder. \crowngen~utilizes these features to predict the deformation from a corresponding point cloud template to the target crown from coarse to fine. The surface of the target crown is generated from the deformed template using Differentiable Poisson Surface Reconstruction (DPSR). A postprocessing algorithm is applied to excise extraneous regions from the reconstructed surface, guided by the cervical margin line detected by \marginseg.

Our main contributions are summarized as follows:
\begin{itemize}

\item We proposed \totalframework, a novel cervical margin aware framework for dental crown design. The framework consists of two learning-based components, \marginseg~and \crowngen. 
\totalframework~improves the generated crowns significantly in both the accuracy and geometric details.

\item We designed \marginseg, a compact network for point cloud segmentation, to detect the cervical margin of the target abutment and impose the location of the cervical margin into \crowngen. The detected cervical margin is also utilized by a tailored post-process to eliminate the superfluous area generated by surface reconstruction, as shown in \autoref{fig:recon}.

\item We introduced a concise network, \crowngen, for crown generation. \crowngen~produces the target crown by deforming and refining the template mesh from coarse to fine. The output of \marginseg~is leveraged to enhance the performance.

\item We established a large-scale intraoral scan dataset from clinical cases and conducted extensive experiments to evaluate the robustness and effectiveness of the proposed method.
\end{itemize}
\section{Related Works}

In this section, we introduce previous studies on the application of artificial intelligence in dental crown design, as well as research related to point cloud segmentation and mesh completion in general computer vision tasks, which can be applied to complete restoration design.

\subsection{Deep Learning Methods for Dental Crown Design}

CAD systems have become ubiquitous in restorative dental crown design, significantly boosting the efficiency of dental technicians. However, compared to other clinical applications, the level of automation and intelligence in current CAD systems for dental crown design remains relatively primitive. Driven by the rapid development of deep learning algorithms, several studies on dental crown generation have emerged in recent years \citep{tian2021dcpr,zhu2022toothcr,hosseinimanesh2025personalized,hosseinimanesh2023mesh,yang2024dcrownformer}.

\cite{tian2021dcpr} addressed dental crown generation by recasting it as a 2D image generation task. They projected IOS data as 2D depth maps from the occlusal perspective and generated the occlusal surfaces of the target crown with a conditional GAN (cGAN)~\citep{isola2017image,goodfellow2020generative}. A series of post-processing steps was subsequently employed to reconstruct the final crown mesh from the generated occlusal surfaces. \cite{zhu2022toothcr} demonstrated the feasibility of dental crown generation using point cloud completion networks by masking a single tooth from IOS data and reconstructing it via a point cloud completion model. \cite{yang2024dcrownformer} proposed a crown generation method based on a Point Transformer architecture. Their model takes IOS point clouds as input and generates the corresponding crown point cloud, which is then converted into a mesh using a “Shape as Points” (SAP) module. However, this method exhibits limited capability in reconstructing fine-grained anatomical details of dental crowns, such as grooves and fossae. \cite{hosseinimanesh2025personalized} utilized AdaPointr \cite{yu2023adapointr}, an advanced point completion network, to generate the dental crown and further modified the loss function to better suit the specific requirements of crown generation. This method demonstrated superior performance compared to similar existing approaches, achieving state-of-the-art results in dental crown generation.

\subsection{Point Cloud Segmentation}

According to the approaches used to process point cloud data, point cloud segmentation algorithms can be categorized into point-wise and voxel-wise methods. PointNet~\citep{qi2017pointnet} and PointNet++~\citep{qi2017pointnet++} are seminal works of the point-wise approach. However, their pooling-based feature aggregation limits the ability to capture the spatial correlation among points. PointTransformer~\citep{zhao2021point} addressed this issue by employing self-attention blocks for adaptive feature learning. However, it requires a large-scale training dataset to achieve satisfactory generalization performance. For the voxel-wise methods, such as MinkowskiNet~\citep{choy20194d}, the point cloud is voxelized as a 3D volume. Networks designed for 3D image segmentation are applied to process this discrete 3D volume. The point cloud segmentation is obtained by reversing the voxelization process to map the volume segmentation back to the point cloud. Voxel-wise methods primarily suffer from two main drawbacks: quantization loss caused by voxelization and substantial computational consumption required by 3D convolutions. Some voxel-wise methods utilize sparse convolutions~\citep{liu2015sparse} to reduce VRAM usage and improve computational efficiency. In recent years, several studies, such as PVCNN~\citep{liu2019point}, have emerged that integrate point-wise and voxel-wise features for point cloud segmentation. These hybrid approaches combine the strengths of both methodologies, demonstrating excellent performance on point cloud segmentation tasks.

\subsection{Point Cloud Completion and Surface Reconstruction}

Due to their unique data structures, existing deep neural networks struggle to process mesh data directly. Consequently, the task of mesh completion is typically divided into two sub-tasks: point cloud completion and surface reconstruction. This decomposition enables the leveraging of deep neural networks for mesh completion. Previous research in point cloud completion has proposed many effective models, including PCN~\citep{yuan2018pcn}, FoldingNet~\citep{yang2018foldingnet}, TopNet~\citep{tchapmi2019topnet}, and GRnet~\citep{xie2020grnet}. However, there is a critical limitation for these methods: the absence of normal vectors for their output points, which are an essential component for most classical mesh reconstruction algorithms. To bridge this gap between point cloud completion and mesh reconstruction, SAP~\citep{peng2021shape} introduced a novel Differentiable Poisson Surface Reconstruction (DPSR) algorithm and integrated it with a point cloud completion network for the direct surface reconstruction from the point cloud. SAP has been proven to be a noise-robust mesh reconstruction approach, capable of reconstructing surface meshes from noisy point clouds. However, similar to traditional surface reconstruction algorithms, such as Poisson Surface Reconstruction (PSR)~\citep{kazhdan2013screened}, SAP generates watertight meshes. This inherent characteristic directly contradicts the desired open, genus-zero morphology of dental crown meshes. This critical discrepancy has received only limited attention in prior research.

\begin{figure}[t]
    \centering
    \includegraphics[width=\linewidth]{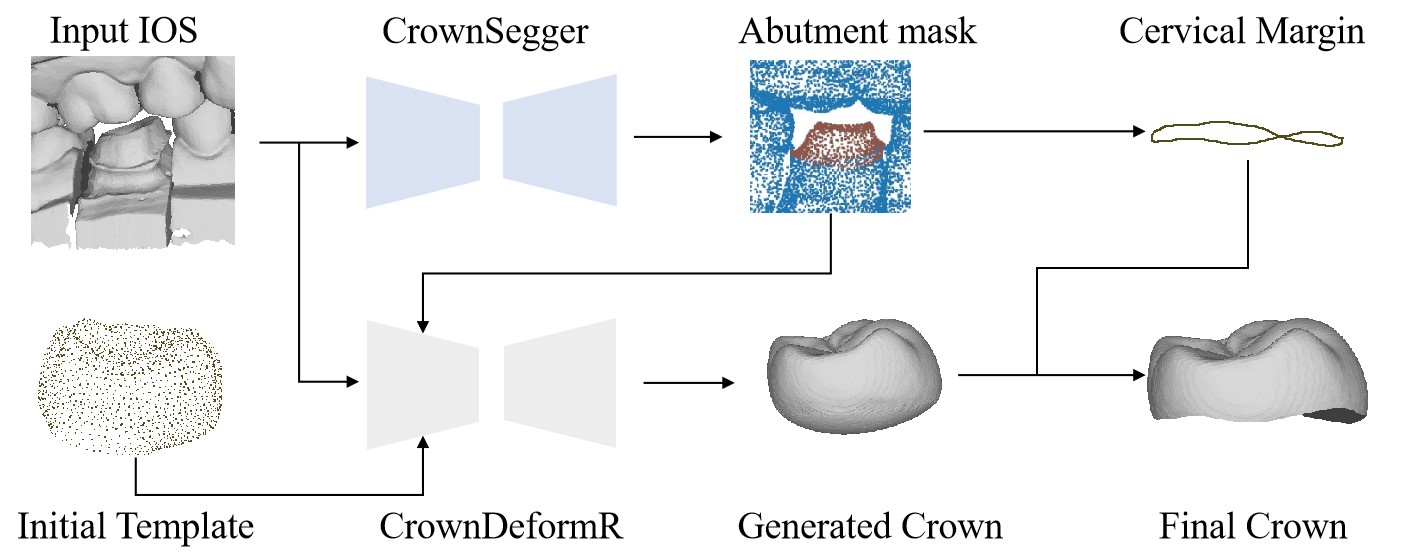}
    \caption{The workflow of \totalframework. The input IOS and the cervical margin identified by \marginseg~are sent to \crowngen~to generate a watertight crown by deforming and refining an initial template. The extraneous regions are excised, guided by the margin line, to obtain the result.}
    \label{fig:workflow}
\end{figure}

\begin{figure*}
    \centering
    \includegraphics[width=\linewidth]{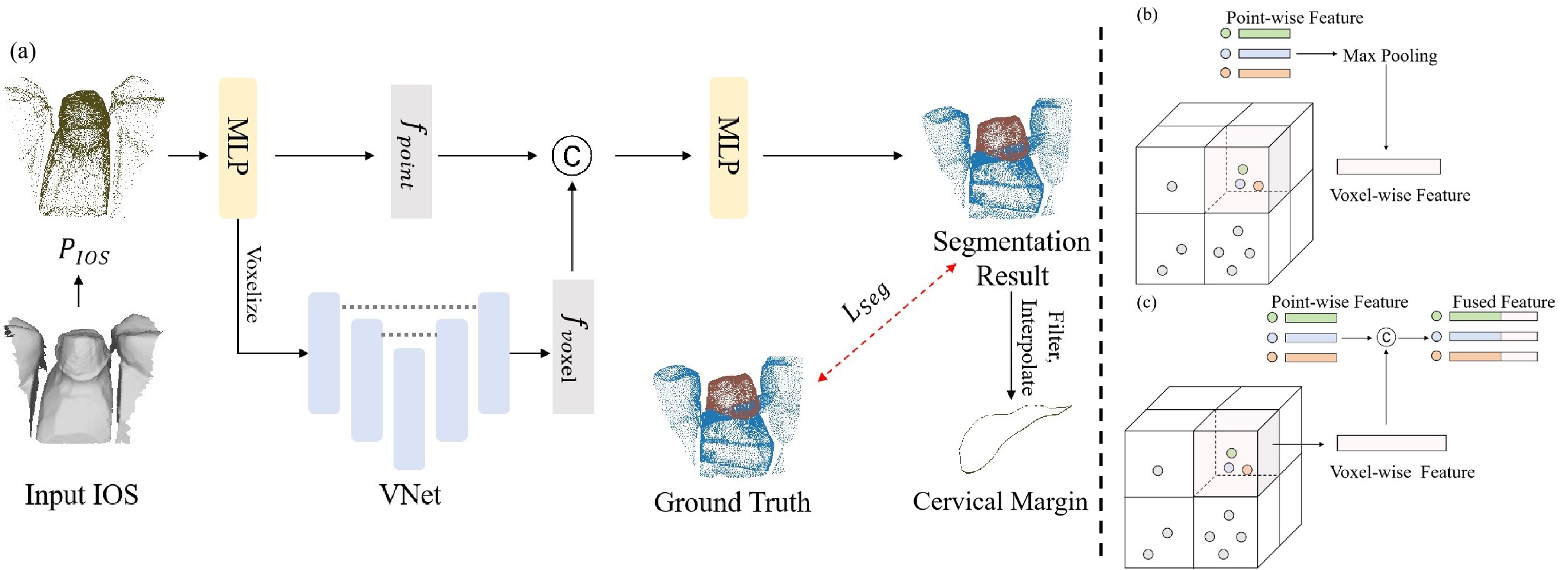}
    \caption{Architecture of \marginseg. The \marginseg~fuses both the point-wise and voxel-wise features and predicts the segmentation of the prepared abutment. The margin line is obtained by extracting the boundary of the segmentation mask.}
    \label{fig:crownsegger}
\end{figure*}

\begin{figure*}[t]
    \centering
    \includegraphics[width=\linewidth]{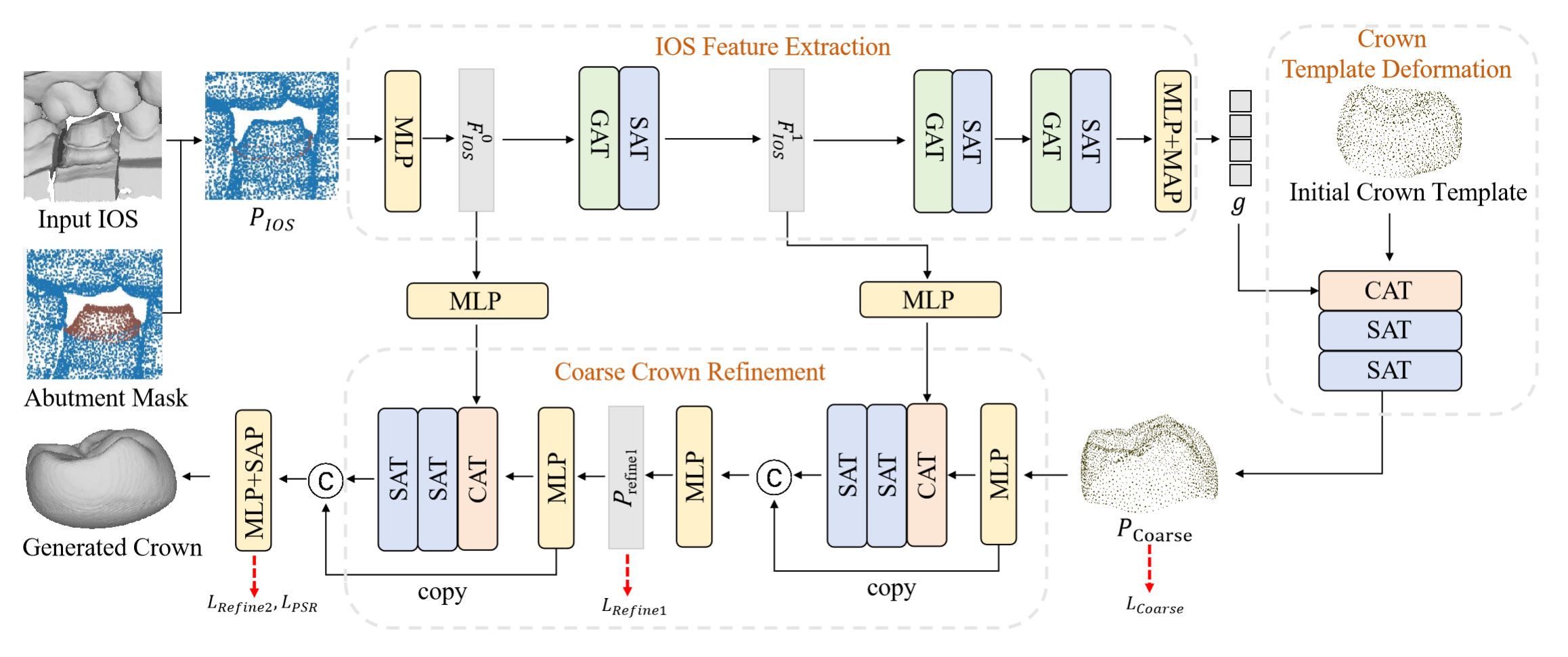}
    \caption{Overall architecture of \crowngen. \crowngen~consists of three components: IOS feature extraction, Crown Template Deformation, and Coarse Crown Refinement. For each generated crown mesh, we utilize the margin line identified by \marginseg~to excise the extraneous regions and solve the intrinsic watertightness issue introduced by DPSR.}
    \label{fig:main}
\end{figure*}

\section{Methods}

The overall pipeline of \totalframework\ is illustrated in \autoref{fig:workflow}. 
Given a target abutment, the cervical margin is first identified by \marginseg\ to extract the cervical margin. 
Conditioned on this margin line and the IOS, \crowngen\ learns to model detailed dental anatomy via attention mechanism. 
By attending to the spatial relationships of adjacent and antagonist teeth, the network adaptively deforms an initial template crown which serves as a geometric prior to facilitate stable training. 
This data-driven deformation produces a crown point cloud that satisfies functional occlusion and proximal contact requirements. 
A watertight crown mesh is subsequently reconstructed using SAP. 
Due to the watertight constraint imposed by SAP, topological artifacts (e.g., spurious mesh extensions near the cervical boundary) may arise. 
A dedicated post-processing module is therefore applied to remove these reconstruction-induced artifacts, ensuring compliance with clinical requirements while preserving the learned anatomical morphology.

\subsection{\marginseg}

The \marginseg, illustrated in \autoref{fig:crownsegger}, is a hybrid point-voxel framework for point cloud segmentation. The initial point-wise features are constructed by concatenating the spatial coordinates and normal vectors. The input point cloud is voxelized into a volume, where each voxel may contain multiple points. Voxel-wise features are then generated by aggregating point-wise features within each voxel through cascaded voxel feature encoding (VFE) layers \citep{zhou2018voxelnet}. These sparse voxel-wise features are transformed into a dense 3D volume, as shown in \autoref{fig:crownsegger} (b), and processed by a VNet~\citep{abdollahi2020vnet} backbone to capture multi-scale anatomical patterns. We concatenate the voxel-wise features from the output feature volume with their corresponding point-wise features, as illustrated in~\autoref{fig:crownsegger} (c), and fuse them with an MLP to predict the segmentation results. The training of \marginseg~is supervised by cross-entropy loss and dice loss. Finally, the cervical margin of the target abutment is identified by extracting and smoothing the boundary of the prepared abutment.

\subsection{CrownDeformR}
\label{sec:crowngen}

The overall architecture of our \crowngen~is illustrated in \autoref{fig:main}. The IOS and the segmentation of the prepared abutment are first encoded to extract geometric and contextual features. Following a CAD-inspired crown design paradigm, \crowngen~generates the target crown by deforming the corresponding tooth template conditioned on the extracted features. The template is progressively refined in a coarse-to-fine manner to produce the final crown geometry.

\subsubsection{Network Primitives}
The proposed architecture consists of three specialized Transformer-based modules: the Geometry Aware Transformer (GAT), the Self Attention Transformer (SAT), and the Cross Attention Transformer (CAT).
The GAT performs hierarchical feature extraction with progressive downsampling while preserving local geometric structures, enabling robust representation of fine-grained dental details.
Within each resolution scale, the SAT is employed to model long-range dependencies within feature tokens, enhancing internal contextual interactions and enabling globally consistent representations across different stages of the framework.
To guide crown generation conditioned on the IOS context, the CAT models the spatial correspondence between crown queries and IOS features, capturing their relative positional dependencies and geometric interactions.
The detailed mathematical formulations and tensor operations of these modules are provided in \autoref{fig:three transformers} and \autoref{fig:refine} of the \nameref{appendix}.

\subsubsection{Context-Aware Feature Extraction}
In the initial feature encoding stage, the IOS point cloud $P_{\text{IOS}}$ is augmented with the abutment segmentation label to provide explicit margin guidance. This hybrid input is processed through a sequence of \textbf{GAT-SAT} blocks, which yields a global context vector $g \in \mathbb{R}^{1 \times 512}$ along with multi-scale intermediate features $F_{\text{IOS}}$. This hierarchical encoding scheme ensures that the subsequent generative stages are informed by both the global dental alignment and localized margin details.

\subsubsection{Hierarchical Template Deformation and Refinement}
Inspired by clinical CAD workflows, \crowngen\ deforms an initial template $T$ selected via the FDI notation system through a two-stage process. During the initial deformation stage, a CAT block captures the spatial dependencies between the template and the global IOS features, guiding the coarse template deformation. The resulting features are further processed by SAT blocks to predict the coarse crown point cloud $P_{\text{Coarse}}$. Subsequently, in the multi-scale refinement stage, this coarse result is progressively upsampled through two refinement modules. Each module leverages CAT to cross-reference localized IOS features, enabling the network to actively learn and reconstruct fine-grained anatomical details, such as occlusal grooves and proximal contact points.

\subsection{Optimization Objectives}

The \crowngen~is optimized using a hierarchical loss function that enforces geometric fidelity from a coarse-to-fine perspective. To ensure the generated crown captures both stable dental proportions and fine-grained anatomical details, we implement a deep supervision strategy across multiple resolutions. Specifically, for the initial deformation and the first refinement stages, we utilize standard Chamfer Distance (CD) to supervise the point cloud at different scales. The \textbf{coarse loss} ($\mathcal{L}_{\text{Coarse}}$) is calculated between the coarse point cloud $P_{\text{Coarse}}$ and a $1/4$ downsampled ground truth $P_{\text{GT}}^{1/4}$ to establish the basic spatial positioning and crown proportions. Subsequently, the \textbf{first refinement loss} ($\mathcal{L}_{\text{Refine1}}$) supervises the $1/2$ downsampled ground truth $P_{\text{GT}}^{1/2}$, ensuring structural consistency as the point density increases. While these global objectives provide a stable optimization trajectory, we utilize the \textbf{Curvature and Margin Penalty Loss (CMPL, $\mathcal{L}_{\text{Refine2}}$)} for the final refinement stage to capture high-frequency anatomical features such as occlusal fossae and ensure precise marginal fit. Unlike vanilla CD, CMPL explicitly prioritizes high-curvature regions and the cervical boundary, formulated as:
\begin{equation}
\begin{aligned}
\mathcal{L}_{\text{Refine2}} = \frac{1}{|\hat{P}_{\text{Crown}}|}\sum_{p \in \hat{P}_{\text{Crown}}} \left( e^{|\kappa(p)|} + \mathbb{I}_{q \in M(P_{\text{GT}})} \right) \min_{q \in P_{\text{GT}}} \|p - q\|_2  \\
+ \frac{1}{|P_{\text{GT}}|} \sum_{q \in P_{\text{GT}}} \left( e^{|\kappa(q)|} + \mathbb{I}_{q \in M(P_{\text{GT}})} \right) \min_{p \in \hat{P}_{\text{Crown}}} \|p - q\|_2,
\end{aligned}
\end{equation}
where $\hat{P}_{\text{Crown}}$ denotes the final predicted point cloud, $\kappa(\cdot)$ represents the normalized curvature, and $M(P_{\text{GT}})$ refers to the set of points on the ground-truth margin line. The exponential term $e^{|\kappa(p)|}$ adaptively increases the penalty in areas with complex anatomy, while the indicator function $\mathbb{I}_{q \in M}$ enforces strict alignment with the prepared abutment. Finally, the surface reconstruction is supervised by an MSE loss ($\mathcal{L}_{\text{DPSR}}$) on the DPSR grid. The total loss is the sum of all aforementioned components. We discuss detailed technical specifications, including the loss function formulations of $\mathcal{L}_{\text{Refine1}}$, $\mathcal{L}_{\text{Coarse}}$, $\mathcal{L}_{\text{DPSR}}$  and specific tensor shape transformations at each stage in the \nameref{appendix}.

\subsection{Post-processing for Reconstructed Mesh}
Most of the surface reconstruction algorithms, including DPSR, produce meshes that are inherently watertight. However, the dental crown is an open genus-zero mesh. It is necessary to eliminate the overextension from the reconstructed mesh to obtain the final result, a step that is rarely mentioned in previous research \citep{hosseinimanesh2025personalized,hosseinimanesh2023mesh,yang2024dcrownformer}. We design a tailored post-processing procedure to address this issue. The cervical margin is identified as the boundary of the prepared abutment, which is predicted by \marginseg. The points on the margin line are smoothed with B-spline kernels. These smoothed points, along with their centroid, define a surface whose normal vectors align with the tooth growth direction. Faces within the reconstructed mesh that are below this surface are then removed. To ensure the mesh boundary precisely conforms to the margin line, points on the boundary of the processed mesh are projected onto the smoothed margin line. Following these post-processing steps, the reconstructed crown surface is transformed into an open genus-zero mesh, which can be manipulated within CAD systems for clinical application.

\section{Experiments and Results}

\begin{algorithm}[ht]
\caption{Cervical Margin Extraction}
\label{alg:margin-extraction}
\textbf{Input:} \\
Mesh: $\mathcal{M} = (\mathcal{V}, \mathcal{F})$ \\
Point-wise labels for the target abutment: $\mathcal{L}: \mathcal{V} \rightarrow \{0, 1\}$
\\
\textbf{Output:} \\
Cervical margin vertex set: $\mathcal{V}_{\text{margin}}$
\\
\textbf{Step 1: Abutment Extraction}\\
Extract faces from $\mathcal{F}$, in which three vertices are labeled as abutment: 
$\mathcal{F}_1 \leftarrow \{f \in \mathcal{F} \mid \forall v \in f, \mathcal{L}(v) = 1\}$.
\\
\textbf{Step 2: Noise Exclusion} \\
Construct a sub-mesh $\mathcal{M}_1 = (\mathcal{V}_1, \mathcal{F}_1)$. \\
Extract the largest connected component $\mathcal{M}_a$ from $\mathcal{M}_1$.
\\
\textbf{Step 3: Boundary Detection} \\
Extract the vertices $\mathcal{V}_b$ in the boundary of $\mathcal{M}_a$.
\\
\textbf{Step 4: Cervical Margin Estimation} \\
Estimate the cervical margin by B-spline interpolation with $\mathcal{V}_b$. \\
Resample 1,000 vertices ($\mathcal{V}_{\text{margin}}$) uniformly on the cervical margin.
\\
\Return{$\mathcal{V}_{\text{margin}}$}
\end{algorithm}

\subsection{Datasets}

\subsubsection{Dataset for prepared abutment segmentation}

We built a dataset for prepared abutment segmentation that consists of 576 IOS data. Each IOS is point-wise annotated, indicating whether individual points belong to the prepared abutment. Among these, 476 scans are allocated for training, and 100 for testing. The cervical margin, which is defined as the boundary of the prepared abutment, is extracted by \autoref{alg:margin-extraction}.

\subsubsection{Dataset for Dental Crown Design}
\begin{table}[htbp]
\centering
\caption{Data Distribution for Dental Crown Design}
\label{tab:crown_data_distribution}
\begin{tabular}{l l c}
\toprule
Jaw Side & Tooth Position & Number of Cases \\
\midrule
\multirow{4}{*}{Upper} 
  & 1st Premolar  & 543 \\
  & 2nd Premolar & 412 \\
  & 1st Molar     & 1096 \\
  & 2nd Molar    & 270 \\
\midrule
\multirow{4}{*}{Lower} 
  & 1st Premolar  & 152 \\
  & 2nd Premolar & 178 \\
  & 1st Molar     & 1452 \\
  & 2nd Molar    & 499 \\
\bottomrule
\end{tabular}
\end{table}

We collect raw IOS data acquired from 4,602 patients with single tooth defects, in which the target teeth are prepared as abutments. For each patient, the corresponding dental crown designed by dental technicians is also collected. The target teeth in the dataset encompass all the premolars and molars. For each IOS data sample, we extract a 2 cm-sided cubic region centered at the centroid of its corresponding crown. The dataset is divided into training, validation, and testing sets with 8:1:1. We utilize stratified sampling among the types of target teeth, ensuring that the distribution of target teeth is consistent across both the training and test sets. To enhance computational efficiency during training, we pre-extract the curvature for each vertex in the IOS data, as well as the boundaries of the IOS data, specifically within the test set.

\subsection{Implementation Details}
\label{sec:implement}

The \marginseg~was implemented with PyTorch and trained on a NVIDIA GTX 3080 Ti with a batch size of 16 for 300 epochs. We used the AdamW optimizer with an initial learning rate of $10^{-4}$ to update the network parameters and applied a cosine learning rate scheduler during training.

The \crowngen~was also implemented with PyTorch and trained for 500 epochs on 4 NVIDIA RTX 4090 GPUs with a batch size of 8. Configurations of the optimizer and learning rate scheduler for \crowngen~are the same as \marginseg. The abutment mask used by \crowngen~is the zero-shot prediction from \marginseg. The employed transformer blocks (GAT, SAT, and CAT) leveraged multi-head attention with 4 heads and 512 hidden channels. The training of \crowngen~is supervised by the combination of the aforementioned loss functions: 
\begin{equation}
    \mathcal{L}_{\text{total}} = \mathcal{L}_{\text{Coarse}}+\mathcal{L}_{\text{Refine1}}+\mathcal{L}_{\text{Refine2}}+\mathcal{L}_{\text{DPSR}}
\end{equation}

The dental crown generation pipeline consists of two distinct models: \marginseg~and \crowngen. During the inference stage, these two models are combined as a unified network, \totalframework, which comprises 37.3 M parameters. \totalframework~requires only 1.1 GB VRAM for the inference of a single sample. The single-inference latency is approximately 600 ms (Segmentation, Generation and Post-processing) on an NVIDIA L20 GPU. This resource frugality makes the \totalframework~an ideal candidate for on-site deployment.

\definecolor{best}{RGB}{220,230,242}
\definecolor{second}{RGB}{220,242,220}
\begin{table}[ht]
\centering
\caption{Comparison experiment on cervical margin segmentation. The best results are highlighted in \colorbox{best}{\textbf{blue}}. The \marginseg~achieves the best performance on the test set. It also exhibits excellent performance under a zero-shot setting on crown-generated preparations, achieving a Hausdorff Distance of 0.545 mm without additional finetuning.}
\label{tab:margin_comparison}
\resizebox{\textwidth}{!}{
\begin{tabular}{lccc}
\toprule
Methods & Accuracy $\uparrow$ & IoU  $\uparrow$ & Hausdorff Distance (mm) $\downarrow$ \\
\midrule
PointNet \citep{qi2017pointnet} & 0.935 & 0.858 & 2.450 \\
PointNet++ \citep{qi2017pointnet++} & 0.956 & 0.901 & 1.720 \\
PointTransformer \citep{zhao2021point} & 0.963 & 0.939 & 1.238 \\
\marginseg & \cellcolor{best}\textbf{0.991} & \cellcolor{best}\textbf{0.972} & \cellcolor{best}\textbf{0.328} \\
\marginseg (zero-shot on crown generation dataset) & -- & -- & \textbf{0.545} \\
\bottomrule
\end{tabular}}

\end{table}

\begin{table*}[ht]
\centering
\caption{Performance comparison of different crown generation methods before mesh reconstruction. The best and second-best results are highlighted in \colorbox{best}{\textbf{blue}} and \colorbox{second}{green}, respectively.}
\label{tab:performance_comparison}
\resizebox{\textwidth}{!}{
\setlength{\tabcolsep}{3pt}
\renewcommand{\arraystretch}{1.2}
\begin{tabular}{lccc ccc ccc ccc}
\toprule
\multirow{2}{*}{\textbf{Methods}} & 
\multicolumn{3}{c}{\textbf{CD-L2 ($\text{mm}^2$)} $\downarrow$} & 
\multicolumn{3}{c}{\textbf{Fidelity Distance ($\text{mm}^2$)} $\downarrow$} & 
\multicolumn{3}{c}{\textbf{Hausdorff Distance (mm)} $\downarrow$} & 
\multicolumn{3}{c}{\textbf{F-score $\uparrow$}} \\
\cmidrule(lr){2-4} \cmidrule(lr){5-7} \cmidrule(lr){8-10} \cmidrule(lr){11-13}
& Premolar & Molar & Overall 
& Premolar & Molar & Overall 
& Premolar & Molar & Overall 
& Premolar & Molar & Overall \\
\midrule
PCN\citep{yuan2018pcn} +SAP 
& 0.271 & 0.318 & 0.305 
& 0.118 &0.148 & 0.140 
& 1.400 & 1.458 & 1.441 
& 0.850 & 0.889 & 0.878 \\

TopNet\citep{tchapmi2019topnet} +SAP  
& 0.354 & 0.409 & 0.394 
&0.164 & 0.200 & 0.190 
& 1.752 & 1.869 & 1.836 
& 0.770 & 0.812 & 0.800 \\

GRnet\citep{xie2020grnet} +SAP 
& 0.230 &0.254 & \cellcolor{second}0.247 
& 0.110 &\cellcolor{second}0.132 & 0.126 
&  \cellcolor{second}1.118 & \cellcolor{second}1.139 &\cellcolor{second}1.134 
& \cellcolor{second}0.874 & \cellcolor{second}0.901 & \cellcolor{second}0.894 \\

DMCv2 \citep{hosseinimanesh2025personalized} 
& \cellcolor{second}0.227 & 0.264 & 0.254 
& \cellcolor{second}0.107 & \cellcolor{second}0.132  & \cellcolor{second}0.125 
& 1.227 & 1.252 & 1.245 
& 0.868 & 0.897 & 0.888 \\

\totalframework
& \cellcolor{best}\textbf{0.176} & \cellcolor{best}\textbf{0.174} & \cellcolor{best}\textbf{0.175} 
& \cellcolor{best}\textbf{0.098} & \cellcolor{best}\textbf{0.092} & \cellcolor{best}\textbf{0.094} 
& \cellcolor{best}\textbf{1.093} & \cellcolor{best}\textbf{0.996} & \cellcolor{best}\textbf{1.027} 
& \cellcolor{best}\textbf{0.903} & \cellcolor{best}\textbf{0.932} & \cellcolor{best}\textbf{0.924} \\
\bottomrule
\end{tabular}}
\end{table*}

\begin{table*}[ht]

\centering
\caption{Quantitative evaluation of the crown meshes generated by different methods. For a fair comparison, all the outputs of these methods were processed using our post-processing pipeline to remove superfluous areas resulting from surface reconstruction. The best results are highlighted in \colorbox{best}{\textbf{blue}}. To demonstrate the effect of the post-processing algorithm, the last row shows the performance of our method without the post-processing algorithm.}
\label{tab:performance_comparison_mesh}
\resizebox{\textwidth}{!}{
\setlength{\tabcolsep}{3pt}
\renewcommand{\arraystretch}{1.2}
\begin{tabular}{lccc ccc ccc ccc}
\toprule
\multirow{2}{*}{\textbf{Methods}} & 
\multicolumn{3}{c}{\textbf{CD-L2 ($\text{mm}^2$)} $\downarrow$} & 
\multicolumn{3}{c}{\textbf{Fidelity Distance ($\text{mm}^2$)} $\downarrow$} & 
\multicolumn{3}{c}{\textbf{Hausdorff Distance (mm)} $\downarrow$} & 
\multicolumn{3}{c}{\textbf{F-score $\uparrow$}} \\
\cmidrule(lr){2-4} \cmidrule(lr){5-7} \cmidrule(lr){8-10} \cmidrule(lr){11-13}
& Premolar & Molar & Overall 
& Premolar & Molar & Overall 
& Premolar & Molar & Overall 
& Premolar & Molar & Overall \\
\midrule

GRnet+SAP \citep{xie2020grnet}
      & 0.212 & 0.210 &
 0.210
      & 0.111 & 0.100 & 0.103
      & 1.212 & 1.307 &
 1.280
      & 0.895 & 0.901 & 0.899 \\
DMCv2 \citep{hosseinimanesh2025personalized}
      & 0.259 & 0.251 &
 0.253
      & 0.137 & 0.123 & 0.127
      & 1.329 & 1.319 &
 1.322
      & 0.868 & 0.875 & 0.873 \\

VBCD \citep{wei2025vbcd}
&0.212&0.207&0.209&0.108&0.109&0.109&1.115&1.163&1.150&0.896&0.915&0.909\\

Diffusion SDF \citep{chou2023diffusion}
&0.228&0.215&0.219&0.116&0.108&0.110&1.216&1.458&1.390&0.875&0.900&0.893\\

\totalframework
& \cellcolor{best}\textbf{0.198} & \cellcolor{best}\textbf{0.180} & \cellcolor{best}\textbf{0.185} 
& \cellcolor{best}\textbf{0.083} & \cellcolor{best}\textbf{0.087} & \cellcolor{best}\textbf{0.086} 
& \cellcolor{best}\textbf{1.096} & \cellcolor{best}\textbf{1.026} & \cellcolor{best}\textbf{1.046} 
& \cellcolor{best}\textbf{0.907} & \cellcolor{best}\textbf{0.920} & \cellcolor{best}\textbf{0.917} \\
\midrule
\totalframework~w/o Postprocessing 
& 0.556 & 1.186 & 1.006
& 0.086 & 0.094 & 0.091
& 3.176 & 4.342 & 4.016
& 0.826 & 0.831& 0.830 \\

\bottomrule
\end{tabular}}
\end{table*}

\begin{figure}
    \centering
    \includegraphics[width=\linewidth]{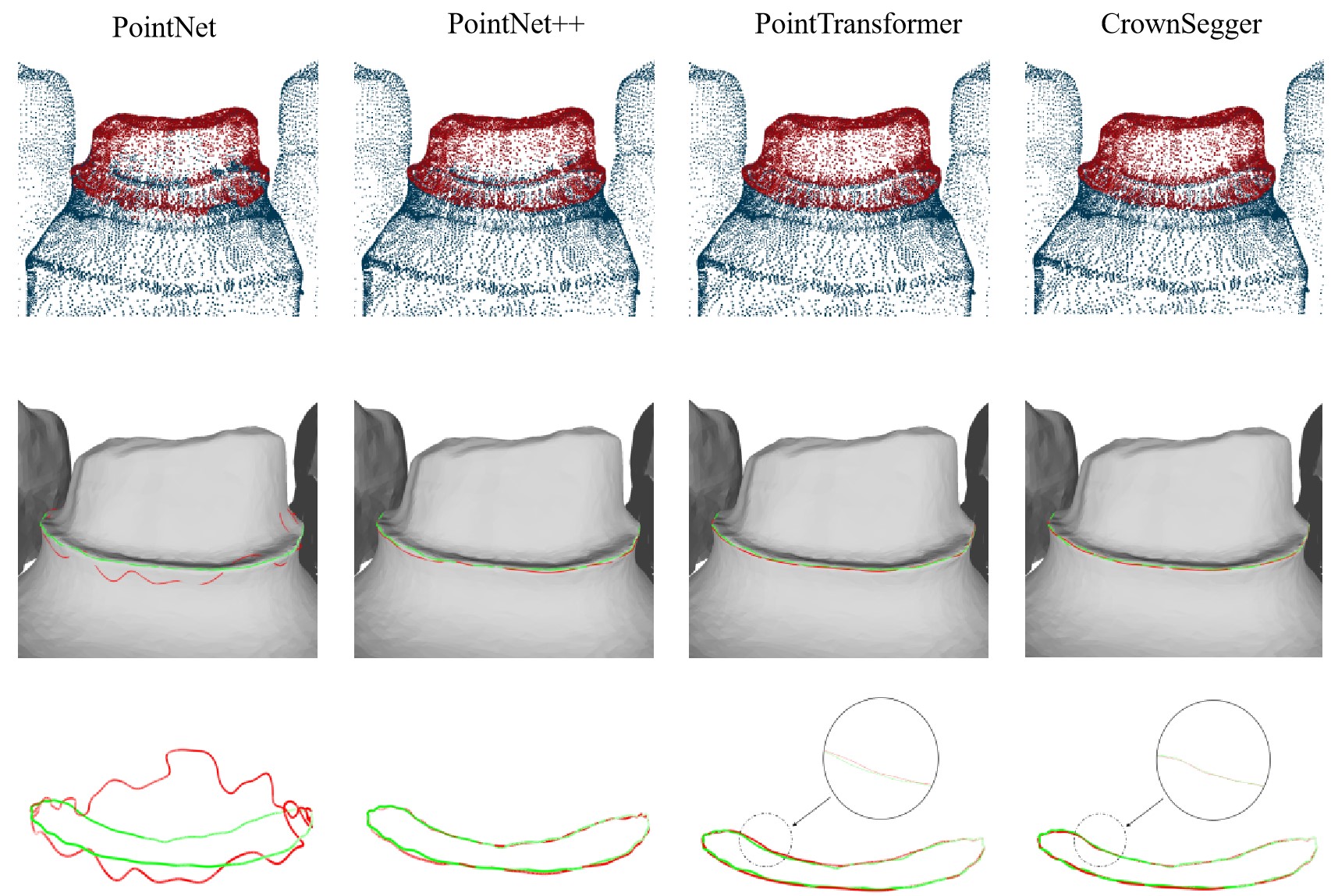}
    \caption{Visualization of the comparison experiment for abutment segmentation. The cervical margin identified by \marginseg~is much closer to the ground truth.}
    \label{fig:Margin_cmp}
\end{figure}

\subsection{Abutment mask Segmentation}
In this section, we conducted a comparative analysis of \marginseg~with several widely adopted point cloud segmentation algorithms, evaluating their performance on both target abutment segmentation and cervical margin extraction. The cervical margin of the target abutment is extracted from its segmentation mask.

For the abutment segmentation, we employed accuracy and Intersection over Union (IoU) as quantitative metrics. The performance of cervical margin extraction was evaluated using the Hausdorff Distance (HDF distance), calculated in the physical coordinate system.

We compared \marginseg~with several prevalent point cloud segmentation models, including PointNet~\citep{qi2017pointnet}, PointNet++~\citep{qi2017pointnet++}, and PointTransformer~\citep{zhao2021point}, on our dataset for abutment segmentation. \marginseg~consistently outperformed these methods across all the evaluated metrics. The improvement in HDF distance is particularly significant, decreasing from 1.237 mm to 0.328 mm (\autoref{tab:margin_comparison}). The performance of cervical margin extraction was also evaluated on the dataset for dental crown design. For this dataset, the ground truth of the cervical margin was defined as the boundary of the dental crown. The cervical margin derived from the zero-shot prediction of \marginseg~achieved remarkable performance with an average HDF distance of 0.575 mm, indicating excellent robustness and generalization ability of \marginseg.

\autoref{fig:Margin_cmp} visualizes an example of the prepared abutment segmentation and the extracted cervical margin. As shown in this figure, \marginseg~demonstrated better segmentation performance, with notably fewer misclassified points. Additionally, the cervical margin extracted from \marginseg's prediction also exhibited better consistency with the ground truth.

It should be noted that \marginseg~is constructed with a transformer-free backbone, which grants it distinct advantages in computational efficiency and memory usage. This makes the model particularly suitable for clinical deployment on edge computing devices, as it achieves high performance with minimal hardware resource demands.
\subsection{Dental Crown Generation}

\subsubsection{Evaluation Metric}

To validate the effectiveness of \totalframework, we conducted extensive comparative experiments against existing dental crown generation methods and ablation studies. We used the L2-normed Chamfer Distance (CD-L2), fidelity distance~\citep{xie2020grnet}, F-score~\citep{hosseinimanesh2023mesh}, and HDF distance for quantitative assessment of experimental results from different perspectives. CD-L2, fidelity distance, and F-score measure the overall similarity between the generated crown and the ground truth. In contrast, the HDF distance quantifies the maximum deviation of the generated crown. HDF distance is a critical evaluation metric for the generated crowns, as it provides an intuitive assessment of their compatibility with the corresponding IOS data. In this section, all distance-related metrics were calculated in the physical coordinate system measured in millimeters (mm).

\subsubsection{Experimental Result}
\label{sec:experiment}
\begin{figure*}[t]
    \centering
    \includegraphics[width=\linewidth]{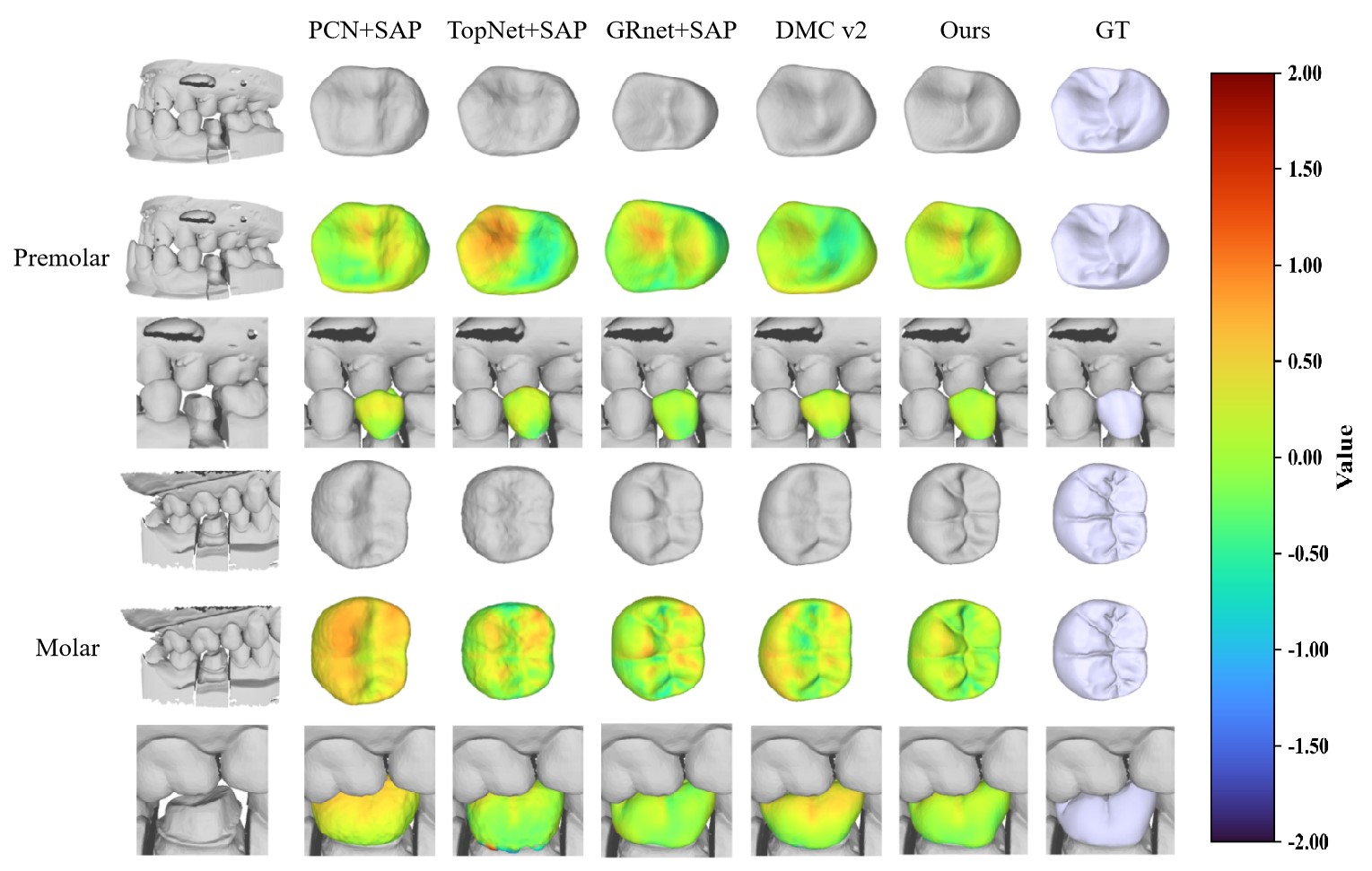}
    \caption{Comparison Result of the \totalframework~with other point cloud based methods.}
    \label{fig:comparison}
\end{figure*}
\begin{figure*}[t]
    \centering
   
    \includegraphics[width=\linewidth]{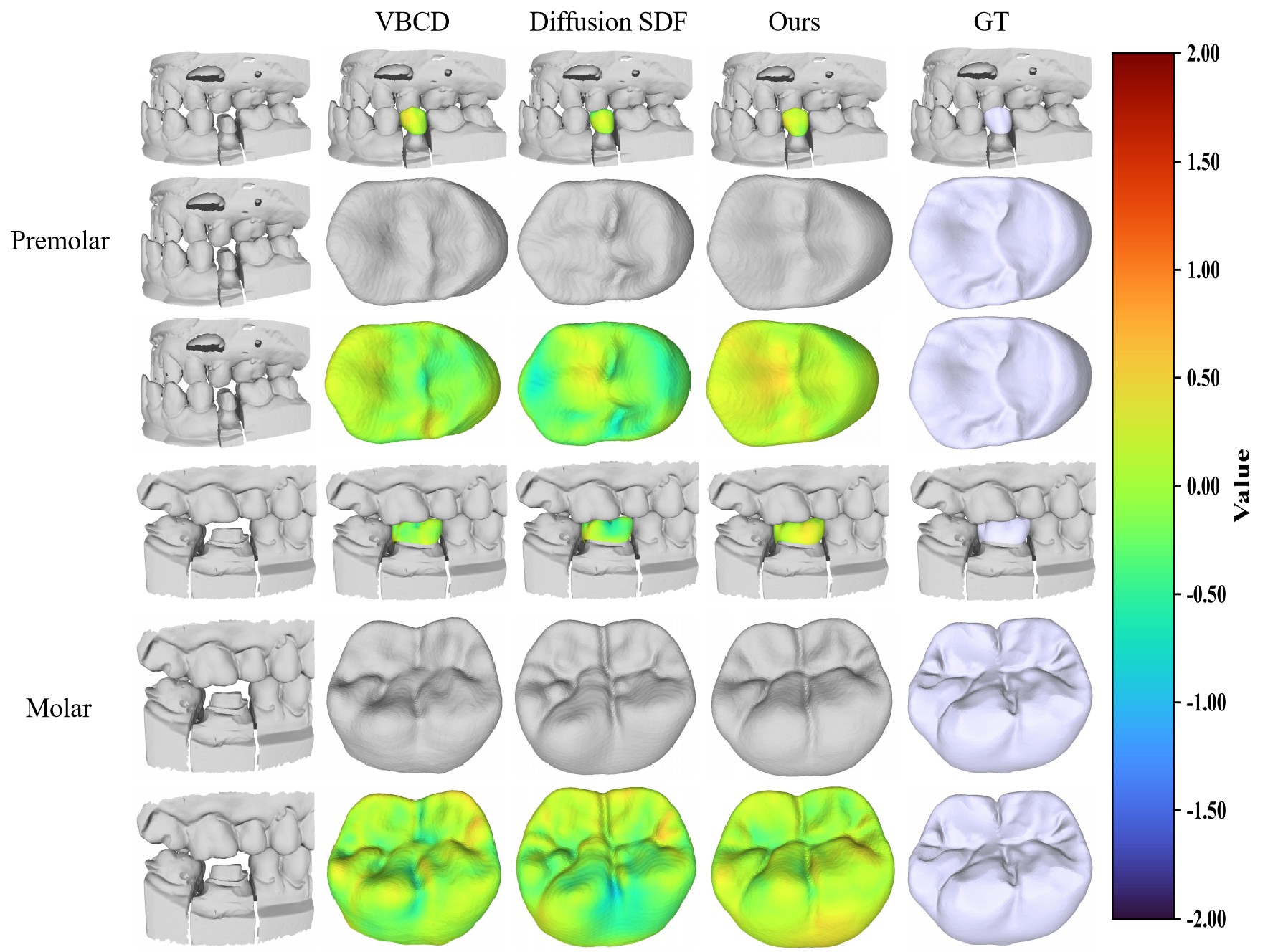}
    \caption{Comparison Result of the \totalframework~with  voxel based methods.}
    \label{fig:comparison_gen}
\end{figure*}

\begin{figure*}
    \centering
    \includegraphics[width=\linewidth]{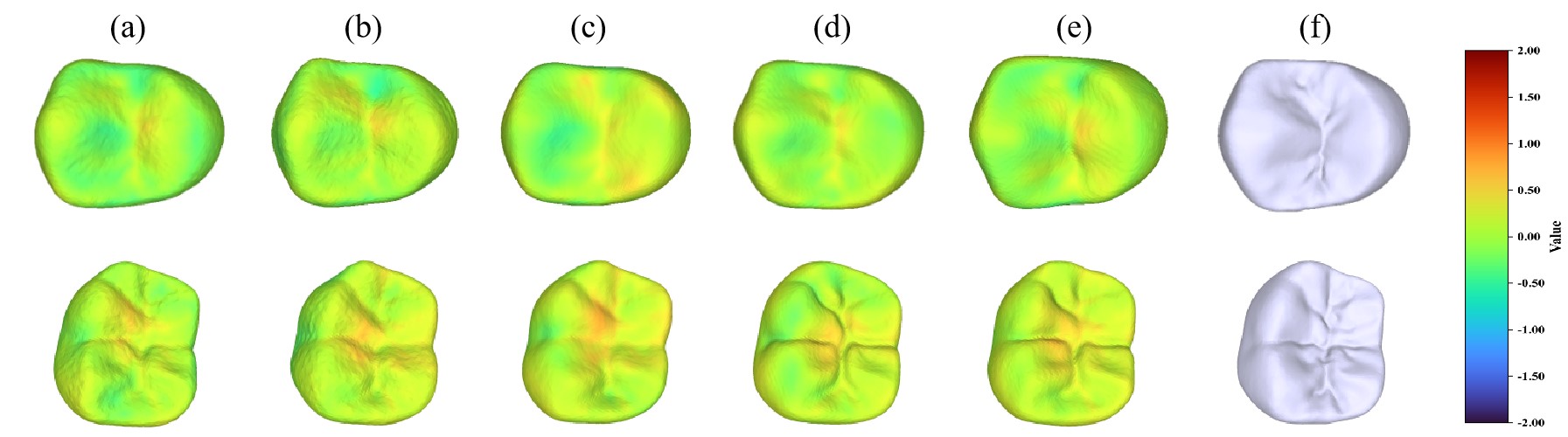}
    \caption{Visualization of the ablation study. (a) to (e) are corresponding to the results in \autoref{tab:ablation}, which are models using: (a) a hemisphere template without the template deformation module; (b) a hemisphere template with the deformation module; (c) the initial crown template without the template deformation module; (d) no cervical margin constraint; and (e) CrownDeformR, our full model. (f) represents the ground truth.}
    \label{fig:ablation study}
\end{figure*}

\begin{figure}
    \centering
    \includegraphics[width=\linewidth]{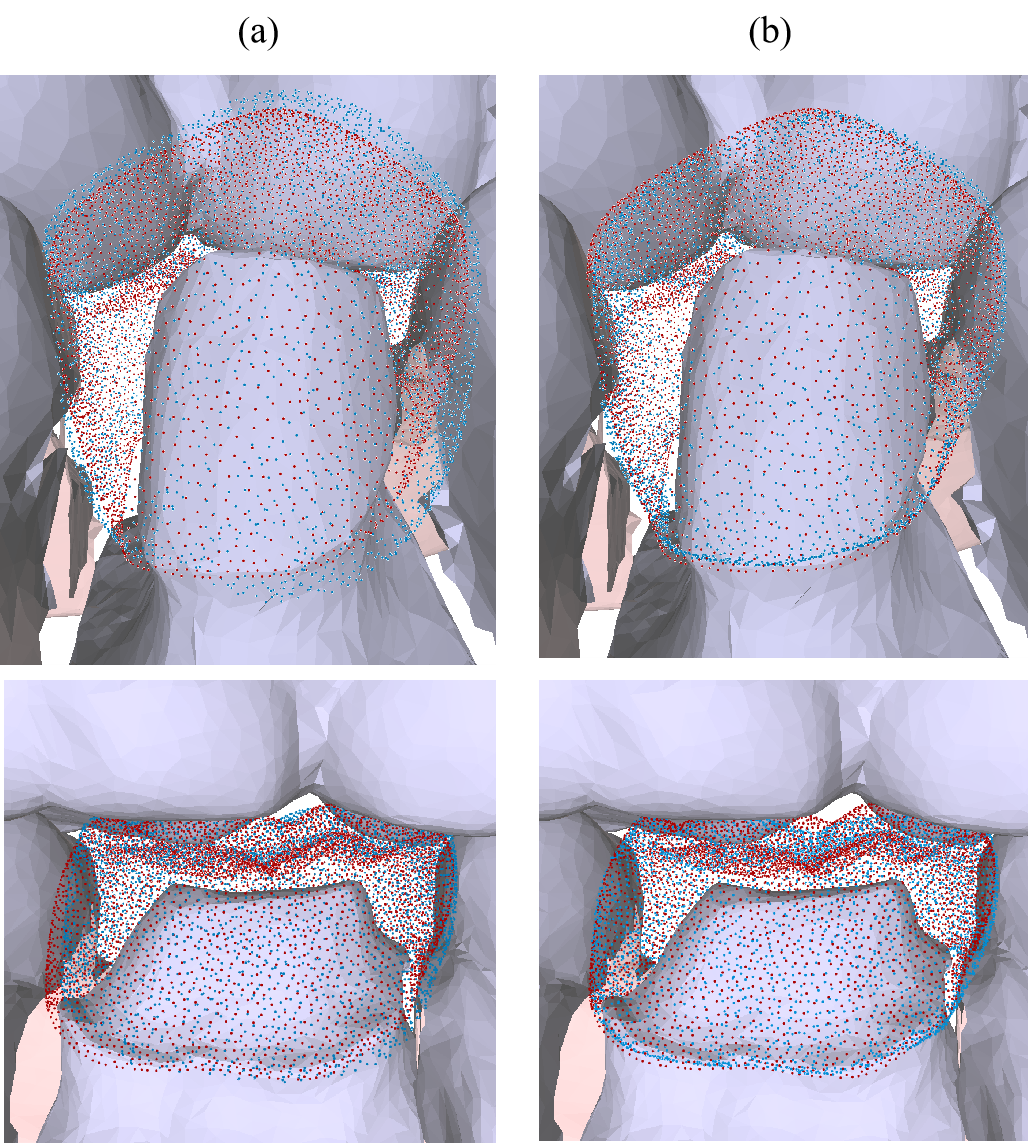}
    \caption{point cloud visualization of the ablation study for cervical margin constraint. The red point cloud denotes the ground truth, and the blue one denotes the generation result. (a) MADCrowner without Cervical Margin Constraint. (b) MADCrowner.}
    \label{fig:margin ablation}
\end{figure}

We compared the performance of \totalframework~with several advanced point cloud completion models, including PCN~\citep{yuan2018pcn}, TopNet~\citep{tchapmi2019topnet}, and GRnet~\citep{xie2020grnet}, which were integrated with SAP~\citep{peng2021shape} to reconstruct the corresponding mesh. Additionally, DMCv2~\citep{hosseinimanesh2025personalized}, the state-of-the-art dental crown generation method, was also evaluated as a comparative experiment.
Following the previous works \citep{hosseinimanesh2023mesh,hosseinimanesh2025personalized}, quantitative metrics were evaluated on the generated point clouds.
The experimental results are listed in \autoref{tab:performance_comparison}. \totalframework~achieved the best performance across all the metrics. Specifically, our method surpassed the second-highest performance in CD-L2, fidelity distance, and F-score by 29.2\%, 26.4\%, and 3.36\%, respectively. This demonstrates that \totalframework~significantly enhanced the consistency between the generated dental crowns and the ground truth. In terms of HDF distance, our method achieved a significant improvement by reducing the best performance from 1.139 mm to 1.027 mm. This notable reduction in HDF distance indicates that our approach generated dental crowns with less noise and lower deviation from the ground truth.
We further conducted a quantitative evaluation for the reconstructed dental crown mesh and incorporated several voxel-based methods \citep{wei2025vbcd, chou2023diffusion} for comparison. The evaluation results are shown in \autoref{tab:performance_comparison_mesh}, where \totalframework~ consistently achieved the best performance across all metrics.
Compared to \autoref{tab:performance_comparison}, there are slight discrepancies in the quantitative metrics of a single method when measured on the generated point cloud versus the reconstructed mesh. These gaps are induced by the inherent perturbations of the mesh reconstruction algorithm and the effect of the postprocessing.

We exhibited visual comparison of \totalframework~to point cloud-based methods and voxel-based methods in
\autoref{fig:comparison} and \autoref{fig:comparison_gen}, respectively. 
The surface distance error of these experimental results is visualized using a colormap. It is obvious that \totalframework~yields the highest consistency between the generated crowns and the ground truth. Additionally, it can be observed that the crowns generated by our method exhibit remarkable improvement in intricate details, such as grooves and fossae.

\subsection{Ablation Study and Cost-Benefit Discussion}
\begin{table*}[ht]
\centering

\caption{
Ablation study results of different components in \crowngen. 
The best results are highlighted in \colorbox{best}{\textbf{blue}}.}
\label{tab:ablation}
\resizebox{\textwidth}{!}{
\begin{tabular}{ccc ccc ccc ccc ccc cc}
\toprule
\multicolumn{3}{c}{\textbf{Components}} & 
\multicolumn{3}{c}{\textbf{CD-L2  ($\text{mm}^2$) $\downarrow$}} & 
\multicolumn{3}{c}{\textbf{Fidelity Distance ($\text{mm}^2$) $\downarrow$}} & 
\multicolumn{3}{c}{\textbf{Hausdorff Distance (mm) $\downarrow$}} & 
\multicolumn{3}{c}{\textbf{F-Score $\uparrow$}} & 
\multicolumn{2}{c}{\textbf{Efficiency}} \\

\cmidrule(lr){4-6} 
\cmidrule(lr){7-9} 
\cmidrule(lr){10-12} 
\cmidrule(lr){13-15}
\cmidrule(lr){16-17}

\makecell{\textbf{Cervical} \\ \textbf{Margin Constraints}} & 
\makecell{\textbf{Initial Crown} \\ \textbf{Template}} & 
\makecell{\textbf{Crown Template} \\ \textbf{Deformation}} & 

Premolar & Molar & Overall & 
Premolar & Molar & Overall & 
Premolar & Molar & Overall & 
Premolar & Molar & Overall & 
\makecell{\textbf{VRAM} (MB) $\downarrow$} & 
\makecell{\textbf{Inference Time} (ms) $\downarrow$} \\

\midrule

\XSolidBrush & \XSolidBrush  & \XSolidBrush & 
0.214 & 0.222 & 0.220 & 
0.106 & 0.114 & 0.112 & 
1.419 & 1.347 & 1.367 & 
0.866 & 0.905 & 0.895 & 
\cellcolor{best}1229 & \cellcolor{best}87 \\

\XSolidBrush & \Checkmark    & \XSolidBrush & 
0.204 & 0.217 & 0.213 & 
0.102 & 0.110 & 0.107 & 
1.372 & 1.291 & 1.314 & 
0.878 & 0.910 & 0.901 & 
1229 & 90 \\

\XSolidBrush & \XSolidBrush  & \Checkmark   & 
0.215 & 0.211 & 0.212 & 
0.115 & 0.106 & 0.108 & 
1.365 & 1.280 & 1.301 & 
0.882 & 0.911 & 0.903 & 
1305 & 100 \\

\XSolidBrush & \Checkmark    & \Checkmark   & 
0.188 & 0.195 & 0.193 & 
0.099 & 0.102 & 0.101 & 
1.182 & 1.152 & 1.160 & 
0.886 & 0.921 & 0.911 & 
1305 & 103 \\

\Checkmark  & \Checkmark     & \Checkmark   & 
\cellcolor{best}\textbf{0.176} &
\cellcolor{best}\textbf{0.174} & 
\cellcolor{best}\textbf{0.175} & 
\cellcolor{best}\textbf{0.098} & 
\cellcolor{best}\textbf{0.092} & 
\cellcolor{best}\textbf{0.094} & 
\cellcolor{best}\textbf{1.093} & 
\cellcolor{best}\textbf{0.996} & 
\cellcolor{best}\textbf{1.027} & 
\cellcolor{best}\textbf{0.903} & 
\cellcolor{best}\textbf{0.932} & 
\cellcolor{best}\textbf{0.924} & 
1305 & 105 \\

\bottomrule
\end{tabular}
}
\end{table*}

The efficacy of \crowngen~hinges upon three critical components: the initial crown template, the template deformation module, and the cervical margin constraints. In this section, we demonstrate the contribution of these components with a series of ablation studies. The results of the ablation studies are listed in \autoref{tab:ablation}. We visualized the results of the ablation studies with an example in \autoref{fig:ablation study}. The dental crown generated by the full \crowngen~demonstrated superior performance in both overall similarity and intricate details compared to the control groups. Beyond the quantitative performance improvements, we further perform a cost-benefit analysis to evaluate the practical necessity of each component. Although the numerical gains introduced by the additional modules may appear modest, such improvements are clinically meaningful in dental crown restoration. Given the high precision requirements of occlusal fitting, even sub-millimeter reductions in geometric deviation can substantially improve patient comfort and reduce the need for chairside adjustments.
Meanwhile, the added computational overhead remains marginal. The incorporation of the template deformation module and cervical margin constraints increases GPU memory consumption (VRAM) by less than 100 MB and inference time by less than 30 milliseconds. Considering this negligible efficiency cost relative to the clinically significant accuracy gains, the complete 
\crowngen
~framework is both practically justified and necessary for high-fidelity dental crown generation.

\subsubsection{Initial crown template}

Inspired by the CAD workflow of dental crown design by technicians, \crowngen~selects an appropriate initial template based on the FDI label of the target tooth. This template is then deformed and refined by \crowngen~to generate the final crown. To evaluate the impact of the specified initial crown template, we conducted an ablation study in which the initial crown template was replaced with a hemisphere of 7.5 mm radius. The results presented in the first and second rows of \autoref{tab:ablation} demonstrate that an appropriate initial crown template can improve the generation results across all the metrics.

\subsubsection{Crown Template Deformation }

In \crowngen, the initial crown template is processed by a template deformation module to create a coarse crown. This coarse representation is then refined to produce the final dental crown. As demonstrated by a comparison of the first and third rows in \autoref{tab:ablation}, the integration of the template deformation module significantly enhances overall crown generation performance compared to the direct utilization of the initial template as an input for the refinement module.

\subsubsection{Cervical Margin Constraints}

The cervical margin serves as a critical reference in dental crown design. In clinical practice, dental technicians begin the crown design by annotating the cervical margin on the prepared abutment. This delineation then guides the subsequent design process, ultimately defining the boundary of the dental crown. \crowngen~also incorporates cervical margin constraints. To enhance awareness of the cervical margin, \crowngen~leverages the prediction of \marginseg~as part of the input, in addition to the IOS data. A comparison of the last two rows in \autoref{tab:ablation} reveals that incorporating the cervical margin constraint significantly enhances all evaluated metrics. Notably, the improvement in HDF distance is particularly pronounced. This is consistent with the fact that the maximum deviation in generated crowns often occurs in the margin region. It suggests that the cervical margin constraint enhances the quality of the generated crown in the critical cervical region. As shown in \autoref{fig:margin ablation}, the application of cervical margin constraints improves the consistency between the generated point cloud and the ground truth. This enhancement is especially notable near the cervical margin.
\subsubsection{Hyper-parameters for CMPL
\label{sec:CMPL}}

\begin{figure}
    \centering
    \includegraphics[width=\linewidth]{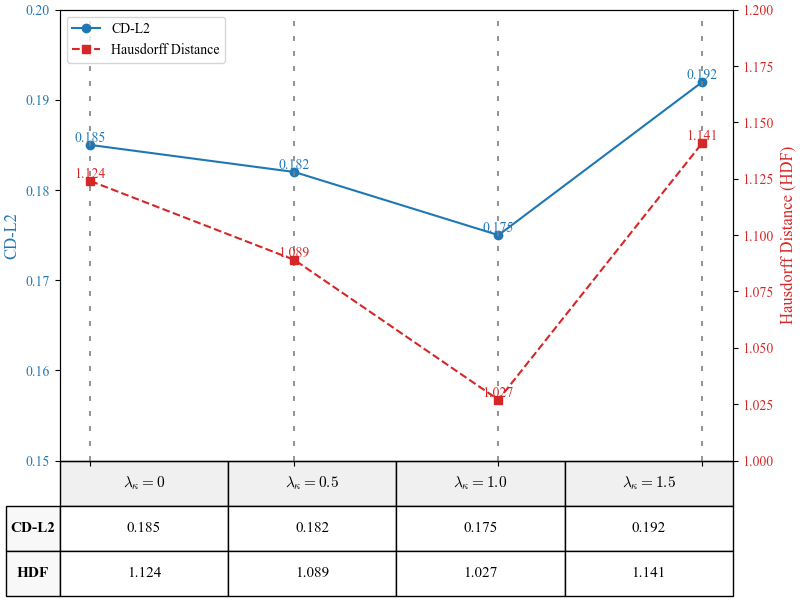}

    \caption{Ablation Study on Curvature Weight.}
    \label{fig:loss_abl}
\end{figure}

As described in the section \nameref{sec:crowngen}, \crowngen~was trained with CMPL to enhance the fidelity of crown details and the peri-cervical regions. CMPL is composed of two distinct parts: \textbf{C}urvature \textbf{P}enalty \textbf{L}oss (CPL) and \textbf{M}argin \textbf{P}enalty \textbf{L}oss (MPL):
$$
\begin{aligned}
\mathcal{CPL} = \frac{1}{|\hat{P}_{\text{Crown}}|}\sum_{p \in \hat{P}_{\text{Crown}}}  e^{\lambda \cdot |\kappa(p)|} \cdot \min_{q \in P_{\text{GT}}} \|p - q\|_2  \\
+ \frac{1}{|P_{\text{GT}}|} \sum_{q \in P_{\text{GT}}} e^{\lambda \cdot |\kappa(q)|} \cdot \min_{p \in \hat{P}_{\text{Crown}}} \|p - q\|_2
\end{aligned}
$$
$$
\begin{aligned}
\mathcal{MPL}=\frac{1}{|\hat{P}_{\text{Crown}}|}\sum_{p \in \hat{P}_{\text{Crown}}}  \mathbb{I}_{q \in M(P_{\text{GT}})} \cdot \min_{q \in P_{\text{GT}}} \|p - q\|_2  \\
+ \frac{1}{|P_{\text{GT}}|} \sum_{q \in P_{\text{GT}}} \mathbb{I}_{q \in M(P_{\text{GT}})} \cdot \min_{p \in \hat{P}_{\text{Crown}}} \|p - q\|_2\\
\end{aligned}
$$
$$
\mathcal{CMPL} = \mathcal{CPL} + \mathcal{MPL}
$$
$\lambda$ is the scale factor of curvature weights. We conducted an ablation study to find the optimal $\lambda$ for dental crown generation. As shown in \autoref{fig:loss_abl}, both CD-L2 and Hausdorff Distance achieved the best performance when $\lambda$ is set to 1. It was adopted as the experimental setting in this work.

\begin{figure}
    \centering

    \includegraphics[width=\linewidth]{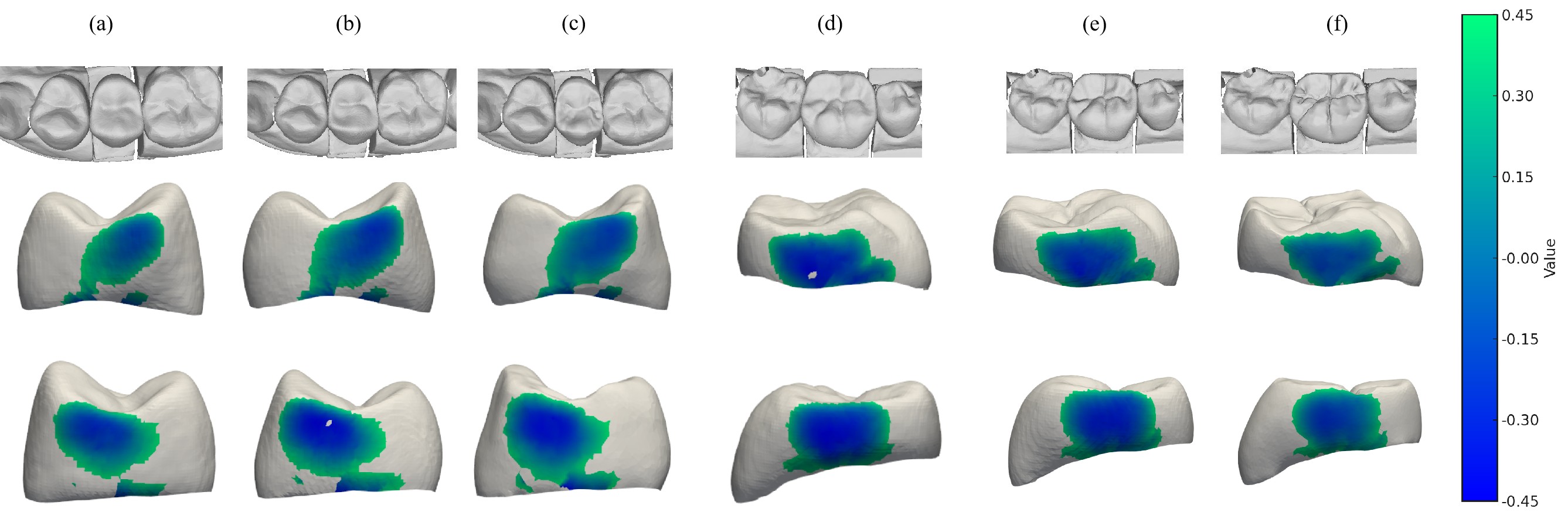}
    \caption{Proximal Contact Visualization. The color map represents the nearest distance to the adjacent teeth. (a) The generation result of a premolar of VBCD. (b) The generation result of a premolar of MADCrowner. (c) Ground truth of the premolar case. (d) The generation result of a molar of VBCD. (e) The generation result of a molar of MADCrowner. (f) Ground truth of the molar case.}
    \label{fig:collision}
\end{figure}

\subsubsection{Mesh Reconstruct Methods}
\label{sec:mesh_reconstruction}

\begin{figure}[h]

    \centering
    \includegraphics[width=\linewidth]{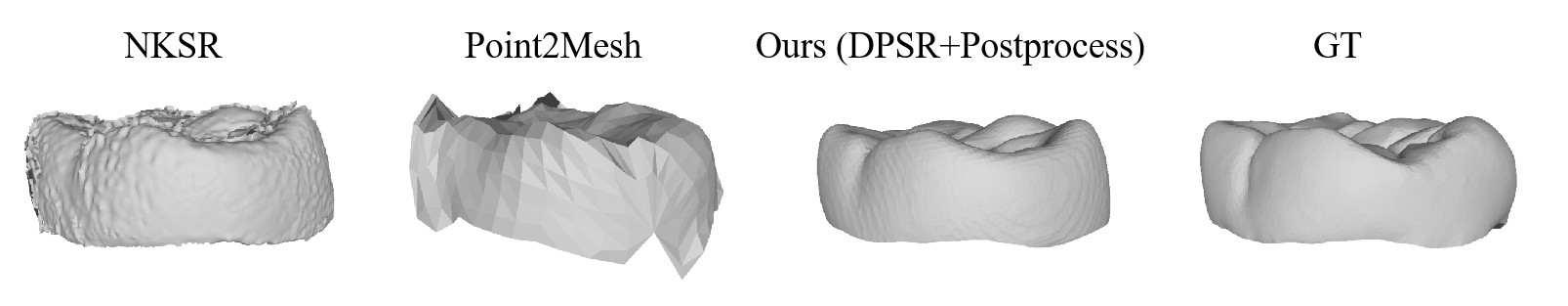}
    \caption{Visualization of different mesh reconstruction method results}
    \label{fig:mesh_reconstruct_cmp}
\end{figure}

In this section, we evaluate the effects of different mesh reconstruction algorithms. For comparative analysis, we utilized Neural Kernel Surface Reconstruction (NKSR) \citep{huang2023neural} and Point2Mesh \citep{hanocka2020point2mesh} to reconstruct surfaces from the point clouds generated by \totalframework. The comparison results are exhibited both visually and quantitatively, in \autoref{fig:mesh_reconstruct_cmp} and \autoref{tab:ablation_reconstruct}. It is evident that the performance of our reconstruction method surpassed other approaches.

\begin{table*}[ht]

\centering\caption{Ablation study on postprocessing and performance comparison of different crown generation methods in \textbf{mesh}. The best results are highlighted in \colorbox{best}{\textbf{blue}}.}
\label{tab:ablation_reconstruct}
\resizebox{\textwidth}{!}{
\setlength{\tabcolsep}{3pt}
\renewcommand{\arraystretch}{1.2}
\begin{tabular}{lccc ccc ccc ccc}
\toprule
\multirow{2}{*}{\textbf{Methods}} & 
\multicolumn{3}{c}{\textbf{CD-L2 ($\text{mm}^2$)} $\downarrow$} & 
\multicolumn{3}{c}{\textbf{Fidelity Distance ($\text{mm}^2$)} $\downarrow$} & 
\multicolumn{3}{c}{\textbf{Hausdorff Distance (mm)} $\downarrow$} & 
\multicolumn{3}{c}{\textbf{F-score $\uparrow$}} \\
\cmidrule(lr){2-4} \cmidrule(lr){5-7} \cmidrule(lr){8-10} \cmidrule(lr){11-13}
& Premolar & Molar & Overall 
& Premolar & Molar & Overall 
& Premolar & Molar & Overall 
& Premolar & Molar & Overall \\
\midrule
MADCrowner (Point2Mesh)
& 0.247 & 0.238 & 
0.240      
& 0.129 & 0.145 & 0.140      
& 1.841 & 1.832 & 
1.834      
& 0.842 & 0.849 & 0.847 \\
MADCrowner (NKSR)
& 0.235 & 0.227 & 
0.229      
& 0.132 & 0.126 & 0.128      
& 1.568 & 1.485 & 
1.508      
& 0.857 & 0.863 & 0.861 \\
\midrule
Ours 
& \cellcolor{best}\textbf{0.198}& \cellcolor{best}\textbf{0.180} & \cellcolor{best}\textbf{0.185}
& \cellcolor{best}\textbf{0.083} & \cellcolor{best}\textbf{0.087}& \cellcolor{best}\textbf{0.086}
& \cellcolor{best}\textbf{1.096}& \cellcolor{best}\textbf{1.026} & \cellcolor{best}\textbf{1.046} 
& \cellcolor{best}\textbf{0.907} & \cellcolor{best}\textbf{0.920}& \cellcolor{best}\textbf{0.917} \\
\bottomrule
\end{tabular}}
\end{table*}

\subsection{Proximal Contact Analysis}

\begin{table*}[ht]
\centering
\caption{Proximal Contact Area deviation of different methods. The best and the second-best results are highlighted in \colorbox{best}{\textbf{blue}} and \colorbox{second}{green}, respectively.}
\label{tab:left_right_ratio}
\resizebox{\textwidth}{!}{
\begin{tabular}{lcc}
\toprule
\textbf{Methods} 
& \textbf{Medial Area Difference ($\text{mm}^2$)} $\downarrow$ 
& \textbf{Lateral Area Difference ($\text{mm}^2$)} $\downarrow$ \\
\midrule
PCN+SAP \citep{yuan2018pcn}          
&  6.51  &  7.30  \\

TopNet+SAP \citep{tchapmi2019topnet} 
&  15.97  &  16.50  \\

GRnet+SAP \citep{xie2020grnet}       
&  5.06  &   4.90  \\

DMCv2 \citep{hosseinimanesh2025personalized}  
&  5.18  &  \cellcolor{second}4.72  \\

VBCD \citep{wei2025vbcd}  
&  \cellcolor{second}5.00 &  {5.04}  \\

Diffusion SDF\citep{chou2023diffusion}  
&  5.37  &  5.18  \\

\totalframework               
&  \cellcolor{best}\textbf{4.37}  &  \cellcolor{best}\textbf{4.07}  \\
\bottomrule
\end{tabular}}
\end{table*}

Appropriate proximal contacts with adjacent teeth are a critical consideration in dental crown design. Excessive contact can lead to periodontal ligament damage, while overly loose contacts may result in food impaction. To address these concerns, dental technicians commonly incorporate appropriate intersection regions between the crown and adjacent teeth. This intentional design provides clinicians with the necessary flexibility to make precise adjustments during crown restoration, ensuring optimal proximal contacts of the target crown. To quantitatively assess the fidelity of proximal contacts, we calculated the proximal intersection area (PIA) between the dental crown and its adjacent teeth. The deviation in PIA between the generated crown and the ground truth then served as the metric for evaluating the accuracy of the proximal contacts for the generated results. \autoref{tab:left_right_ratio} shows that the results of \totalframework~exhibit the lowest error, which indicates that the dental crowns produced by \totalframework~maintain excellent consistency with the ground truth in proximal contacts. 
As depicted in \autoref{fig:collision}, the proximal contact regions of the generated crowns exhibit excellent fidelity with the ground truth in both premolars and molars. Compared with VBCD, the model with the strong geometric performance in \autoref{tab:performance_comparison_mesh}, our method produces proximal contact distributions that are more consistent with the ground truth in terms of spatial localization and contact extent.

\subsection{Failure Case Analysis}
\label{sec:failure}
\begin{figure*}
    \centering

    \includegraphics[width=\linewidth]{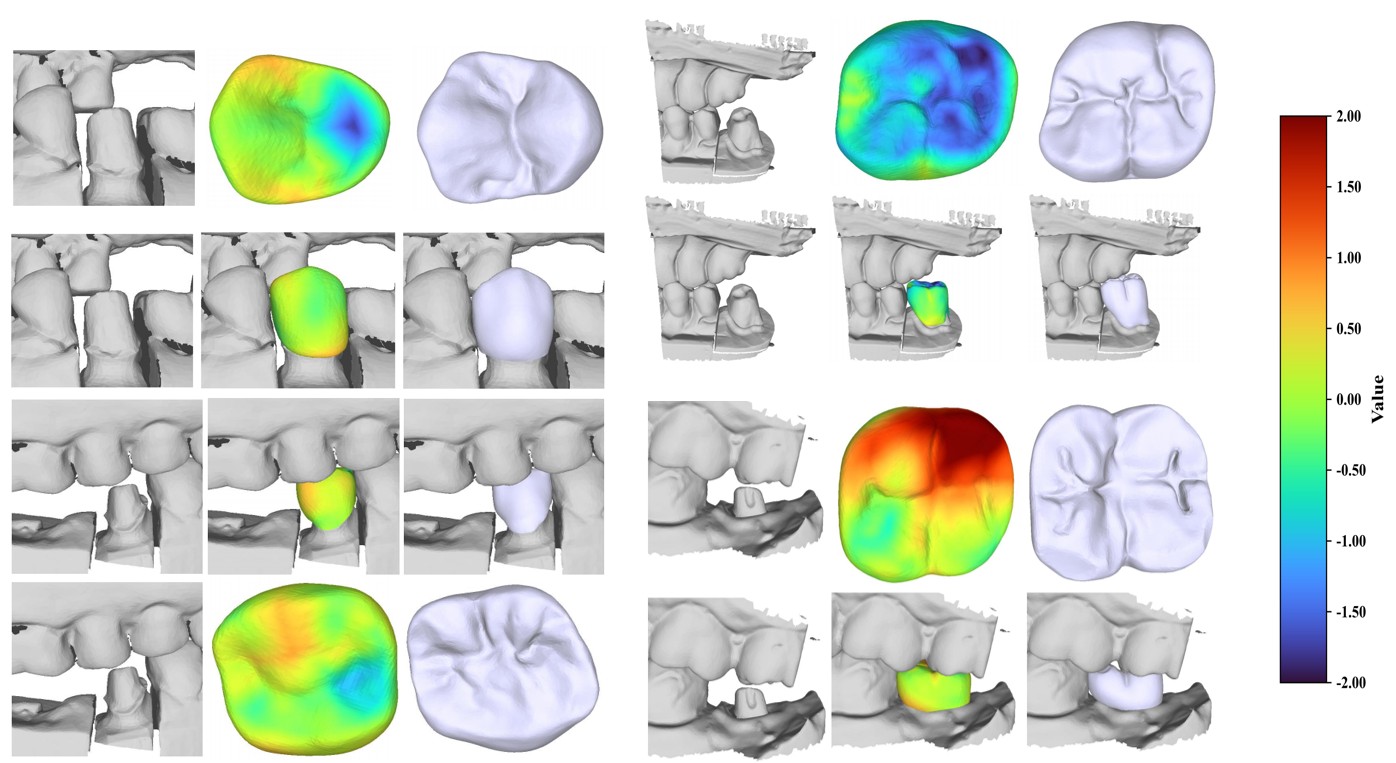}
    \caption{Failure Cases Demonstration}
    \label{fig:placeholder}
\end{figure*}

The \totalframework~ demonstrated strong generalization ability and produced high-quality crown meshes for the majority of cases in the test set. However, beyond isolated extreme outliers, we observed several recurring failure patterns associated with specific clinical conditions.
\autoref{fig:placeholder} presents representative failure cases generated by MADCrowner. Our analysis reveals that crown generation is more prone to failure under the following conditions: (1) inadequate tooth preparation, where irregular or insufficient reduction compromises geometric inference; (2) incomplete intraoral scans, particularly holes or missing regions in adjacent teeth; and (3) absence of adjacent or antagonist teeth, which limits accurate reconstruction of occlusal relationships. In these scenarios, the generated crowns may exhibit improper occlusal contacts or suboptimal morphological adaptation.
The identified patterns are consistent with clinical experience, highlighting the importance of standardized tooth preparation and complete adjacent and antagonist tooth information for reliable automated crown design.

\begin{figure}
    \centering
    \includegraphics[width=\linewidth]{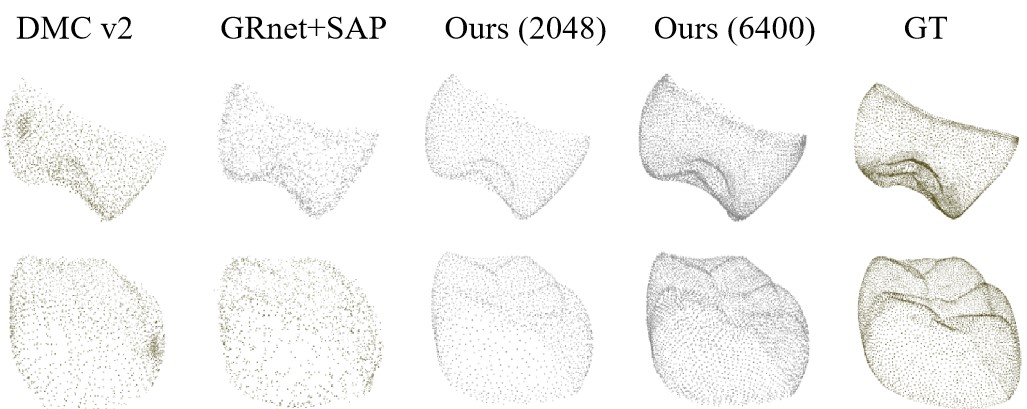}
    \caption{The point cloud result from three models: DMCv2,GRnet+SAP,\totalframework. To facilitate clearer visualization of the noisy point cloud, the predicted point cloud is downsampled to 2048 points using the farthest point sampler.}
    \label{fig:pc_comparison}
\end{figure}

\begin{figure}
    \centering
    \includegraphics[width=1\linewidth]{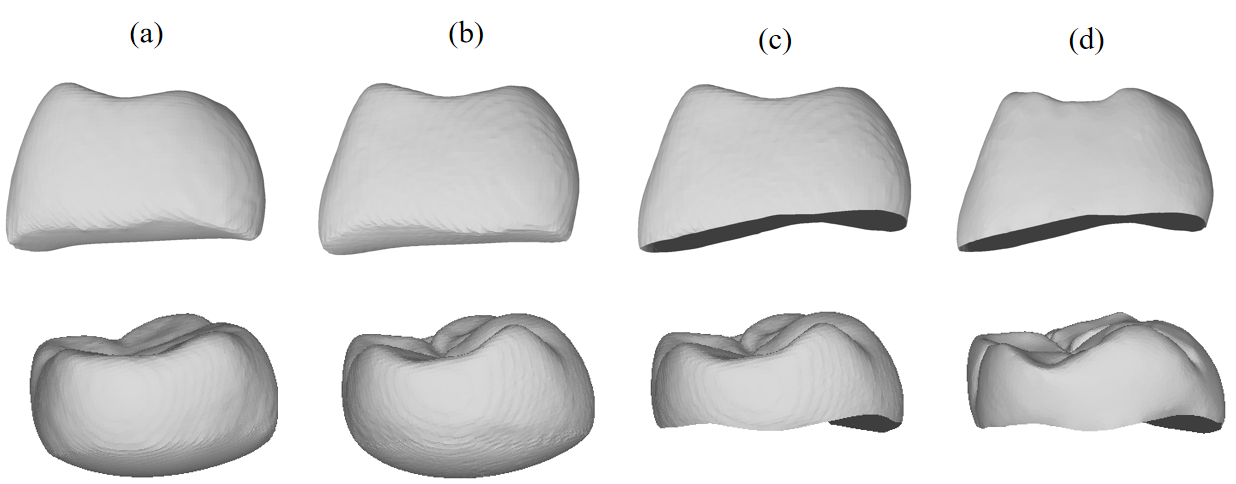}
    \caption{Side view of dental crowns. (a) The generated crown of the previous method, such as DMC. (b) The generated crown of \totalframework~without post-processing. (c) The generated crown of \totalframework. (d) Ground Truth. The inherent watertight issue introduced by DPSR is addressed by our post-processing method.}
    \label{fig:post-processing}
\end{figure}

\begin{figure}[h]
    \centering

    \includegraphics[width=\linewidth]{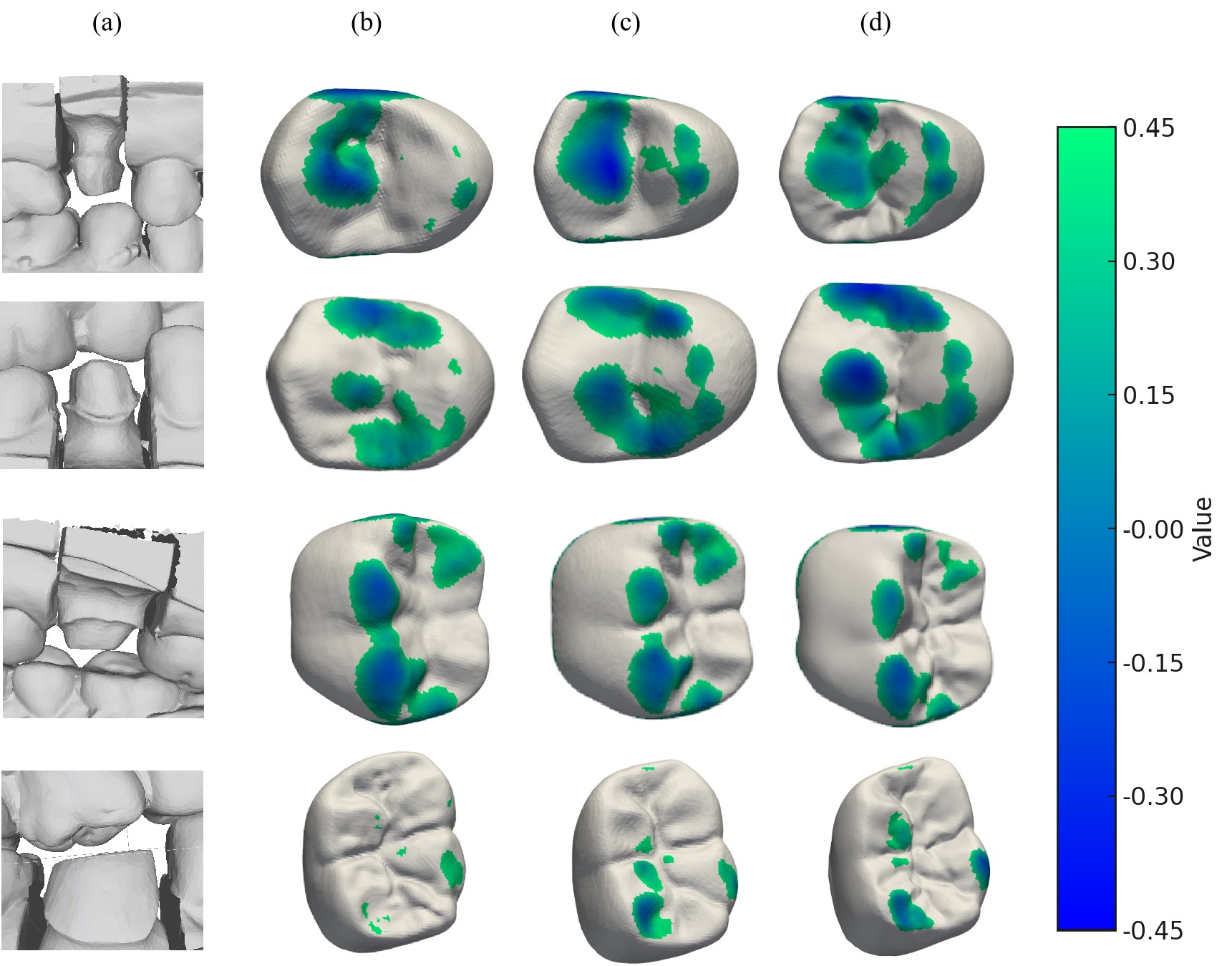}
    \caption{Occlusal analysis. (a) Input IOS, (b) Generated crown of VBCD (c) Generated crown of MADCrowner, (d) Ground truth. For each vertex in the crown, the signed nearest distance to the IOS is visualized with a color map.}
    \label{fig:occlusal}
\end{figure} 
\section{Discussion and Conclusion}

In this work, we propose a novel dental crown generation framework, \totalframework, which primarily consists of \marginseg~and \crowngen. The IOS data is first processed using \marginseg~to generate a segmentation mask of the prepared abutment, from which the cervical margin is subsequently extracted. \crowngen~integrates the IOS data and the abutment mask to generate the dental crown for the target tooth. Our approach achieved superior performance across all the evaluated metrics, including CD-L2, Fidelity Distance, HDF Distance, and F-score. Compared to the approximately 15 minutes required for a dental technician to manually design a crown using a CAD system, \totalframework~generates a dental crown within 500 ms. This dramatic acceleration significantly enhances the efficiency of dental technicians.

For dental crown generation methods based on point cloud completion networks, a uniform spatial distribution and sufficient density of the output point cloud are crucial for accurate dental crown generation. \autoref{fig:pc_comparison} compares the point clouds generated by DMC v2, GRnet, and \totalframework. Both DMC v2 and GRnet fail to generate the anatomical details of the occlusal surface, such as grooves and fossae. GRnet voxelizes the point cloud, which inherently limits the output point cloud density due to the spatial resolution constraints of the input volume. DMC v2 addresses this insufficient point cloud density by directly generating point proxies from IOS data using a point transformer encoder. However, the encoder of DMC v2 and similar methods~\cite{wang2024pointattn,yu2021pointr,yu2023adapointr} is built upon self-attention blocks. All the features are derived exclusively from the IOS data. This often leads to an inhomogeneous spatial distribution of the generated point cloud. To solve this limitation, \crowngen~employed~cross-attention blocks to generate point proxies, which integrate features from the IOS data with the spatial information from the specific initial crown template. As shown in \autoref{fig:pc_comparison}, \crowngen~generates point clouds with a uniform spatial distribution and excellent occlusal surface details.

It should be noted that meshes produced by most of surface reconstruction algorithms are inherently watertight. This property introduces overextension in the reconstructed crown, a challenge that has rarely been discussed in previous studies. As illustrated in \autoref{fig:post-processing}, \totalframework~addresses this issue by incorporating a post-processing method that trims these extraneous areas with reference to the cervical margin. In addition, the post-processing algorithm also guarantees accurate consistency between the margins of the generated crown and the ground truth.

Besides proximal contacts, appropriate occlusal contacts are also crucial for dental crown design. \totalframework~produces dental crowns with favorable occlusal contacts. As depicted in \autoref{fig:occlusal}, which visualizes the occlusal contacts for two typical crowns, the distribution of occlusal contact regions on the generated crowns reveals great similarity with the ground truth. Compared with VBCD, a method that achieves the best performance in \autoref{tab:performance_comparison_mesh}, our method produces occlusal contact regions whose spatial distribution and contact intensity patterns more closely resemble those of the ground truth.

While our experimental results demonstrate that \totalframework~outperforms existing methods, certain limitations persist. Dental crown generation is an inherently open-ended task. Dental technicians often design noticeably different crowns for the same target tooth, reflecting the diverse range of clinically acceptable solutions. However, current methodologies, including ours, typically constrain the predicted crown shape through regression-based losses, such as the Chamfer Distance. This approach inadvertently encourages the network to produce a “smoothed average” of plausible outputs rather than capturing the true “mode” of the underlying distribution. Consequently, the generated crowns may lack critical anatomical details, such as grooves and fossae, which are essential for optimal function and aesthetics.

Future work should explore the incorporation of advanced generative models, such as diffusion models~\citep{ho2020denoising}, within 3D crown generation to produce high-quality crowns with richer morphological details. Moreover, expansion of the IOS dataset is also crucial to further enhance and validate the generalization capability of crown generation algorithms. This expansion should particularly focus on cases in which the incisor or canine is the target tooth.

\section*{CRediT authorship contribution statement}
\textbf{Linda Wei:}Methodology, Validation, Visualization, Writing $-$ original draft. 
\textbf{Chang Liu:}Methodology, Writing $-$ original draft.
\textbf{Wenran Zhang:}Writing $-$ original
draft, Visualization.
\textbf{Yuxuan Hu:}Writing $-$ original draft.
\textbf{Ruiyang Li:}Writing $-$ original draft.
\textbf{Feng Qi:} Writing $-$ original draft.
\textbf{Changyao Tian:}Writing $-$ original draft. 
\textbf{Ke Wang:}Writing $-$ original draft. 
\textbf{Yuanyuan Wang:}Project administration, Supervision, Writing $-$ review \& editing.
\textbf{Shaoting Zhang:}Project administration, Supervision, Writing $-$ review \& editing.
\textbf{Dimitris Metaxas:}Project administration, Supervision
\textbf{Hongsheng Li:} Project administration, Supervision, Writing $-$ review \& editing, Funding Acquisition

\section*{Data availability}
Due to privacy issues, the data used in this study is confidential.

\section*{Declaration of generative AI and AI-assisted technologies in the writing process.}
During the preparation of this work, the authors used ChatGPT in order to correct grammatical errors. After using this tool, the authors reviewed and edited the content as needed and take full responsibility for the content of the published article.

\section*{Declaration of competing interest}The authors declare that they have no known competing financial interests or personal relationships that could have appeared to influence the work reported in this paper.
\section*{Acknowledgement}
The authors would like to thank The Chinese University of Hong Kong, Fudan University, SenseTime, and Rutgers University for providing computational resources and methodology guidance. They also wish to express their appreciation to the Department of Second Dental Center at Shanghai Ninth People’s Hospital, Shanghai Jiao Tong University School of Medicine, and Shanghai Stomatological Hospital for their professional insights and support throughout the project. Finally, the authors are grateful to the Centre for Perceptual and Interactive Intelligence (CPII) under InnoHK and the Natural Science Foundation of Sichuan Province for their funding support.

\section*{Funding}
This project was funded in part by the Centre for Perceptual and Interactive Intelligence (CPII) Ltd under the Innovation and Technology Commission (ITC)’s InnoHK initiative, and in part by the Guangdong Basic and Applied Basic Research Foundation (Grant No. 2023B1515130008, XW).
This work was also supported in part by the National Natural Science Foundation of China (Grant No. 62271115), and in part by the Natural Science Foundation of Sichuan Province (Grant No. 2025ZNSFSC0455).

\bibliographystyle{elsarticle-harv} 
\bibliography{bibtex.bib}
\clearpage

\section*{Appendix}
\label{appendix}

\subsection*{A. Details of The Network Design}
\subsubsection*{A.a Three Transformers}
\label{sec:detail of design}
\begin{figure}[!h]
    \centering
    \includegraphics[width=\linewidth]{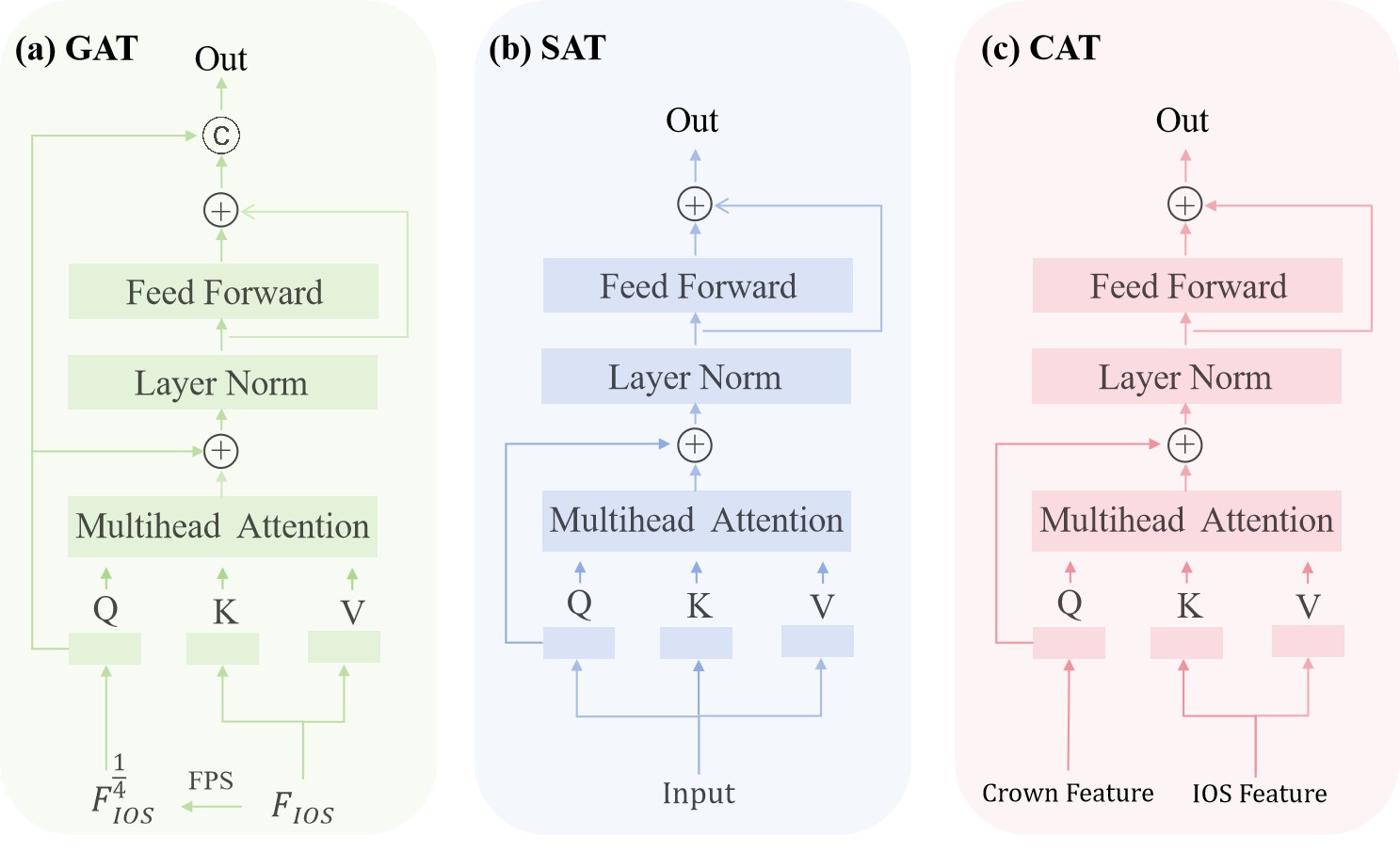}
    \caption{Detailed structures of a Geometry Aware Transformer (GAT), a Self Attention Transformer (SAT), and a Cross Attention Transformer (CAT)}
    \label{fig:three transformers}
\end{figure}

There are three attention blocks mentioned in the section \nameref{sec:crowngen}. The detailed structures of these blocks are illustrated in \autoref{fig:three transformers}.

The key and value of GAT are generated from IOS feature vectors denoted as $F_{\text{IOS}}$, while the query is generated from $F_{\text{IOS}}^{\frac{1}{4}}$, which is obtained by applying farthest point sampling to $F_{\text{IOS}}$, reducing the number of feature vectors to $1/4$ based on their spatial distribution. Both global and local geometric details of the point-wise feature can be captured by GAT.

\begin{figure}
    \centering
    \includegraphics[width=\linewidth]{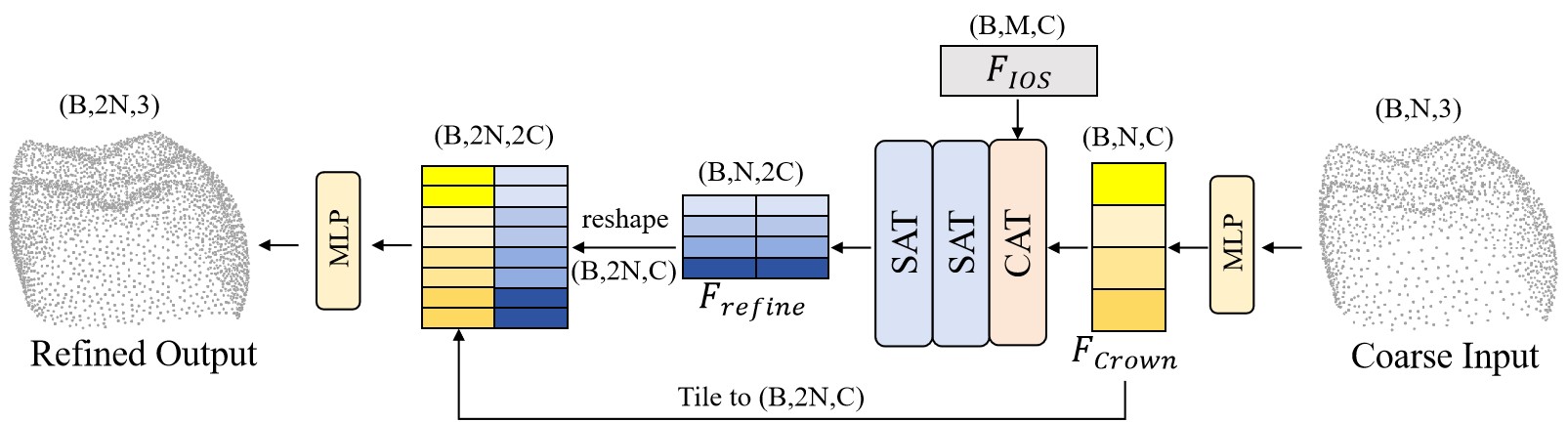}
    \caption{Details of the Coarse Crown Refinement module. The number of points in the generated crown doubled after the processing.}
    \label{fig:refine}
\end{figure}

\subsubsection*{A.b Detail of CrownDeformeR}
\noindent\textbf{IOS Feature Extraction and Cervical Margin Constraints.} 
In personalized dental crown design, the position and morphology of the adjacent and opposing teeth are crucial references to ensure proper functional restoration and aesthetics. Additionally, the cervical margin of the target abutment serves as an essential guideline for crown design. For a successful restoration, the crown boundary must accurately conform to the cervical margin. Therefore, we merge the segmentation of the prepared abutment into the input of \crowngen~to enhance the cervical margin constraint. The input of \crowngen~is denoted as $P_{\text{IOS}} \in \mathbb{R}^{n\times 4}$, where the first three dimensions represent the 3D coordinates of the IOS point cloud and the fourth dimension represents the segmentation label. We first encode $P_{\text{IOS}}$ to feature vectors $F_{\text{IOS}}^{0} \in \mathbb{R}^{n\times 64}$ using an MLP. $F_{\text{IOS}}^{0}$ is then processed by a sequence of Geometry Aware Transformer (GAT) and Self Attention Transformer (SAT)~\citep{wang2024pointattn}. GAT is a variant of SAT, in which the number of output feature vectors is reduced to $\frac{1}{4}$ of the input. The output of the GAT-SAT sequence is aggregated as the global IOS feature $g \in \mathbb{R}^{1 \times 512}$ by an MLP and a max pooling layer. $F_{\text{IOS}}^{0}$ and the intermediate features $F_{\text{IOS}}^{1} \in \mathbb{R}^{\frac{n}{4} \times 128}$ after the first GAT-SAT block are utilized to provide context features at different scales for the refinement of the coarse crown in the subsequent stage.

\noindent\textbf{Crown Template Deformation.}
Inspired by the procedure of CAD dental crown design, \crowngen~deforms a template crown to generate the personalized crown. In a typical adult dentition, there are 28 teeth, each annotated with a unique label according to the Fédération Dentaire Internationale (FDI) notation system~\citep{international1984dentistry}. In \totalframework, we assign a point cloud crown template for each tooth. The appropriate template based on the target tooth is selected as the initial crown $T \in \mathbb{R}^{m\times 3}$. \crowngen~deforms the initial crown $T$ by incorporating the global feature $g$ through a Cross Attention Transformer (CAT) block. We use an MLP to encode $T$ as feature vectors $F_{\text{template}} \in \mathbb{R}^{m \times 128}$, which is utilized to generate the query of the CAT block, while the key and value are obtained from the global IOS feature $g$. The output of CAT is further processed by two SAT blocks to generate the coarse point cloud of the personalized crown, denoted as $P_{\text{Coarse}}^{0}$. Instead of directly utilizing the global IOS feature to generate all the points~\citep{yang2024dcrownformer}, the CAT block facilitates interaction between the global IOS feature and the initial point cloud, leading to more effective and specific generation results. The deformation of the crown template is supervised by the chamfer distance loss formulated as:
$$
\begin{aligned}
\mathcal{L}_{\text{Coarse}} &=  \frac{1}{|P_{\text{Coarse}}|} \sum_{p \in P_{\text{Coarse}}} \min_{q \in P_{\text{GT}}^{\frac{1}{4}}} \| p - q \|_2^2 \\
&\quad + \frac{1}{|P_{\text{GT}}^{\frac{1}{4}}|} \sum_{q \in P_{\text{GT}}^{\frac{1}{4}}} \min_{p \in P_{\text{Coarse}}} \| p - q \|_2^2~,
\end{aligned}
$$
where \( P_{\text{GT}}^{\frac{1}{4}} \) refers to a downsampled point cloud of the ground truth crown \( P_{\text{GT}} \), in which the number of points decreases to $\frac{1}{4}$.

\noindent\textbf{Coarse Crown Refinement.} 
The \crowngen~further refines the coarse crown to generate the fine-grained result with a multi-scale paradigm. For each scale, the number of points in the crown point cloud increased by twice after the processing of the crown refinement module, which consists of a sequence of CAT and SAT. We use the CAT block to fuse the IOS feature with the coarse crown. The query of the CAT is derived from the feature vectors $F_{\text{Crown}} \in \mathbb{R}^{B \times N \times C}$, which is obtained from the coarse crown with an MLP. The corresponding IOS features $F_{\text{IOS}}$ are utilized to generate the key and value. The output of the CAT is further processed by two SAT blocks to generate $F_{\text{refine}} \in \mathbb{R}^{B \times 2 C \times N}$. It is then reshaped as $\mathbb{R}^{B \times C \times 2N}$ to generate the refined point cloud. The detailed structure of the coarse crown refinement module is demonstrated in \autoref{fig:refine}.
There are two crown refinement modules in \crowngen. We perform a deep supervision strategy to optimize these modules. For the first module, we use the chamfer distance loss formulated as:
$$
\begin{aligned}
\mathcal{L}_{\text{Refine1}} &=  \frac{1}{|P_{\text{Refine1}}|} \sum_{p \in P_{\text{Refine1}}} \min_{q \in P_{\text{GT}}^{\frac{1}{2}}} \| p - q \|_2^2 \\
&\quad + \frac{1}{|P_{\text{GT}}^{\frac{1}{2}}|} \sum_{q \in P_{\text{GT}}^{\frac{1}{2}}} \min_{p \in P_{\text{Refine1}}} \| p - q \|_2^2~,
\end{aligned}
$$
in which $P_{\text{GT}}^{\frac{1}{2}}$ refers to a $\frac{1}{2}$ down-sample of the $P_{\text{GT}}$, $P_{\text{Refine1}}$ is the refined point cloud.
For the second crown refinement module, we employ \textbf{C}urvature and \textbf{M}argin \textbf{P}enalty \textbf{L}oss (CMPL) to provide enhanced supervision for the crown details and the peri-cervical margin regions. CMPL ensures the crown refinement with adequate details and improved alignment to the cervical margin. CMPL is defined as $\mathcal{L}_{\text{Refine2}}$ in the following formula:
$$
\begin{aligned}
\mathcal{L}_{\text{Refine2}} = \frac{1}{|\hat{P}_{\text{Crown}}|}\sum_{p \in \hat{P}_{\text{Crown}}} \left( e^{|\kappa(p)|} + \mathbb{I}_{q \in M(P_{\text{GT}})} \right) \min_{q \in P_{\text{GT}}} \|p - q\|_2  \\
+ \frac{1}{|P_{\text{GT}}|} \sum_{q \in P_{\text{GT}}} \left( e^{|\kappa(q)|} + \mathbb{I}_{q \in M(P_{\text{GT}})} \right) \min_{p \in \hat{P}_{\text{Crown}}} \|p - q\|_2~,
\end{aligned}
$$
where $\hat{P}_{\text{Crown}}$ is the predicted point cloud after the second refinement module. $\kappa(\mathbf{\cdot})$ is the normalized curvature. $M(P_{\text{GT}})$ is the set of margin line points in the ground truth.
The $\hat{P}_{\text{Crown}}$ is sent to a SAP block to predict the DPSR grid for surface reconstruction. The prediction of the DPSR grid $G_\text{pred}$ is supervised by an MSE loss with the ground truth $G_\text{GT}$:
$$
\mathcal{L}_{\text{DPSR}} = ||G_\text{pred} - G_{\text{GT}}||_2
$$

\subsection*{B. Generalization on Incisors and Canines}
\label{sup:incisor and canine}

Molars and premolars serve as the primary teeth during mastication. In clinical cases, damage to molars and premolars occurs much more frequently than to canines and incisors \citep{wayman1994relative, scavo2011frequency, kunwar2021endodontic}. Accordingly, previous studies on dental crown generation algorithms also primarily focused on these posterior teeth \citep{yang2024dcrownformer, hosseinimanesh2023mesh, hosseinimanesh2025personalized}. In our collected dataset, the number of damaged canines and incisors is substantially lower than that of molars and premolars. The distribution of the raw dataset is shown in \autoref{tab:crown_data_distribution_all}. Given this imbalance, the experiments in the manuscript restricted target teeth to molars and premolars.

\begin{figure}[t]
    \centering
    
    \includegraphics[width=\linewidth]{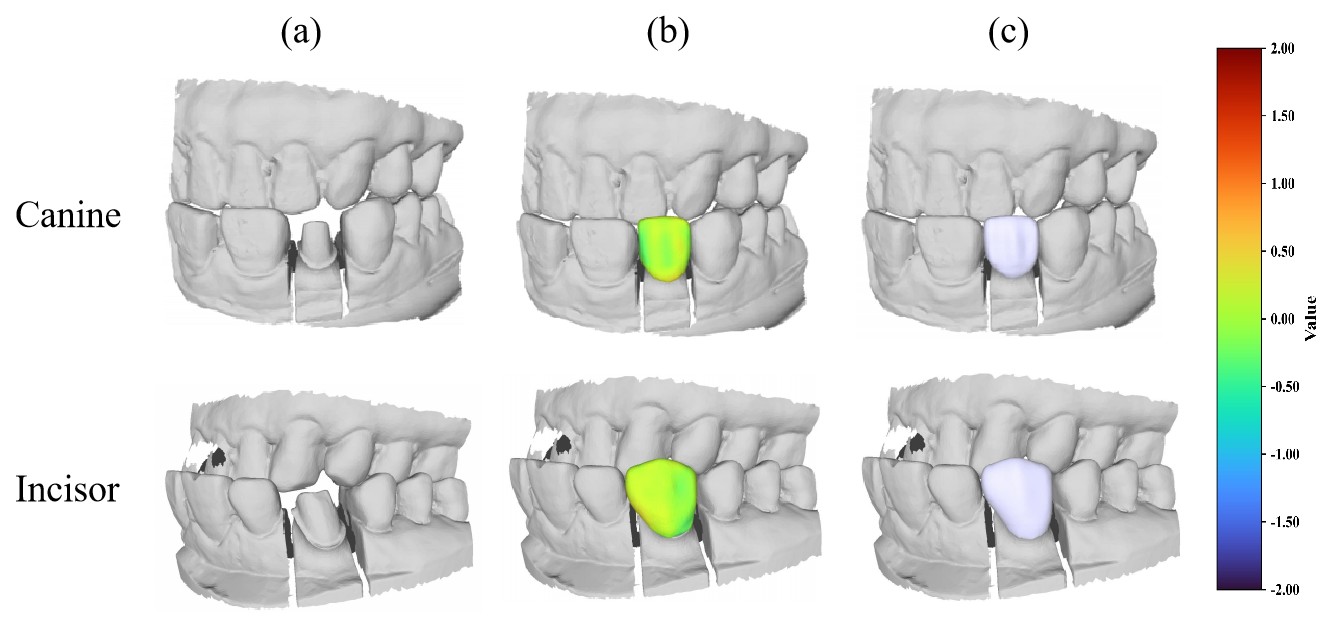}
    \caption{Visualization for the generated dental crown on incisors and canines. (a) Input IOS mesh. (b) Generation results of the retrained model. (c) Ground truth.}
    \label{fig:incisor_canine}
\end{figure}

To demonstrate the generalization capability of the proposed method, we incorporated data samples with canines and incisors as target teeth from the raw dataset into the training and testing sets and retrained our model. \autoref{fig:incisor_canine} visualizes two typical results generated by the retrained model. These results confirm that our method is effective in generating dental crowns for both canines and incisors. For a more comprehensive assessment of the retrained model, we conducted a quantitative evaluation of the generated crowns across different tooth types, and summarized these results in \autoref{tab:all_tooth_types}. Due to the pronounced data imbalance, the generation quality for incisors and canines still exhibits a measurable performance gap when compared to the molars and premolars. In future work, we plan to mitigate this disparity by collecting additional clinical data from patients with damaged canines and incisors.

\begin{table}[htbp]

\centering

\caption{Data distribution of the raw dataset for dental crown design.}
\label{tab:crown_data_distribution_all}
\centering
\begin{tabular}{cccc}
\toprule
Tooth Position & \multicolumn{3}{c}{Number of Cases} \\
& Maxillary & Mandible & Total\\
\midrule
Incisor & 217 & 33 & 250 \\
Canine & 115 & 13 & 128\\ 
Premolar & 955 & 330 & 1285\\
Molar & 1366 & 1951 & 3317\\
\bottomrule
\end{tabular}
\end{table}

\begin{table}[ht]

\centering

\caption{Quantitative evaluation for dental crown generation across different tooth types.}
\label{tab:all_tooth_types}
\setlength{\tabcolsep}{8pt}
\renewcommand{\arraystretch}{1.1}
\resizebox{\textwidth}{!}{
\begin{tabular}{lcccc}
\toprule
Tooth Type & CD-L2 $\downarrow$ & Fidelity Distance $\uparrow$ & F-value $\uparrow$ & Hausdorff Distance (mm) $\downarrow$ \\
\midrule
Incisor  & 0.285 & 0.143 & 0.766 & 1.589 \\
Canine   & 0.406 & 0.174 & 0.731 & 1.820 \\
Premolar & 0.199 & 0.112 & 0.881 & 1.243 \\
Molar    & 0.204 & 0.106 & 0.915 & 1.112 \\
\bottomrule
\end{tabular}}
\end{table}






\end{document}